\documentclass[10pt,twocolumn,letterpaper]{article}

\usepackage[pagenumbers]{cvpr} % To force page numbers, e.g. for an arXiv version
\usepackage[accsupp]{axessibility}

\usepackage{booktabs}
\usepackage{amsfonts}
\usepackage{nicefrac}
\usepackage{microtype}
\usepackage{xcolor}
\usepackage{graphicx}
\usepackage{amsmath}
\usepackage{amssymb}
\usepackage{array}
\usepackage{multirow}
\usepackage{color}
\usepackage{colortbl}
\usepackage{framed}
\usepackage{bm}
\usepackage{xspace}
\usepackage{enumitem}
\usepackage{xparse}
\usepackage{algorithm}
\usepackage{listings}
\usepackage{url}
\usepackage{mathtools}
\usepackage{pifont}
\usepackage{mathtools}
\usepackage{lipsum}
\usepackage{adjustbox}
\usepackage[most]{tcolorbox}
\usepackage{xcolor}
\usepackage{bbm}

\newcommand{\cmark}{\ding{51}}
\newcommand{\xmark}{\ding{55}}

\definecolor{Gray}{gray}{0.2}
\definecolor{lightgray}{gray}{0.92}
\definecolor{blond}{rgb}{0.98, 0.94, 0.75}
\definecolor{TitleColor}{gray}{0.95}
\definecolor{LightCyan}{rgb}{0.88,0.95,1}
\definecolor{OurColor}{rgb}{0.855, 0.937, 0.957}

\definecolor{blond}{rgb}{0.98, 0.94, 0.75}
\def \ie {\emph{i.e.}}
\def \eg {\emph{e.g.}}

\definecolor{customgray}{gray}{0.35}

\newcommand{\tit}[1]{\smallbreak\noindent\textbf{#1.}}
\newcommand{\tinytit}[1]{\noindent\textbf{#1.}}

\newcommand{\ours}{ReAG\xspace}

\newcommand{\inc}[1]{\textcolor{blue!60!black}{\textbf{\small$\Delta$+#1}}}

\definecolor{pastelblue}{rgb}{0.85, 0.95, 0.98}
\definecolor{pastelpink}{rgb}{1.0,0.82,0.86}
\definecolor{pastelgreen}{rgb}{0.47,0.87,0.47}
\definecolor{pastelyellow}{rgb}{0.99,0.99,0.59}

\newtcolorbox{promptbox1}[1][]{
  enhanced,
  breakable,
  colback=pastelblue!20,
  colframe=pastelblue!20!black,
  leftrule=1.5mm,
  arc=1mm,
  boxrule=0.6pt,
  top=2mm, bottom=2mm, left=3mm, right=3mm,
  fonttitle=\bfseries,
  title=Prompt,
  fontupper=\ttfamily,
  #1,
  before upper={\color{black}},
}

\definecolor{promptbox2bg}{HTML}{EEF6E9}

\newtcolorbox{promptbox2}[1][]{
  enhanced,
  breakable,
  colback=promptbox2bg,
  colframe=promptbox2bg!20!black,
  leftrule=1.5mm,
  arc=1mm,
  boxrule=0.6pt,
  top=2mm, bottom=2mm, left=3mm, right=3mm,
  fonttitle=\bfseries,
  title=Prompt,
  fontupper=\ttfamily,
  #1,
  before upper={\color{black}},
}

\definecolor{promptbox3bg}{rgb}{0.992, 0.945, 0.965}
\definecolor{text3}{rgb}{0.15,0.15,0.15}

\newtcolorbox{promptbox3}[1][]{
  enhanced,
  breakable,
  colback=promptbox3bg,
  colframe=promptbox3bg!20!black,
  leftrule=1.5mm,
  arc=1mm,
  boxrule=0.6pt,
  top=2mm, bottom=2mm, left=3mm, right=3mm,
  fonttitle=\bfseries,
  title=Prompt,
  fontupper=\ttfamily,
  #1,
  before upper={\color{text3}},
}

\definecolor{placeholdercolor}{rgb}{0.1,0.3,0.6}

\newcommand{\placeholder}[1]{\textcolor{placeholdercolor}{\texttt{\{#1\}}}}

\newcommand\blfootnote[1]{%
  \begingroup
  \renewcommand\thefootnote{}\footnote{#1}%
  \addtocounter{footnote}{-1}%
  \endgroup
}

\definecolor{cvprblue}{rgb}{0.21,0.49,0.74}
\usepackage[pagebackref,breaklinks,colorlinks,allcolors=cvprblue]{hyperref}

\def\confName{CVPR}
\def\confYear{2026}

\title{\ours: Reasoning-Augmented Generation for Knowledge-based\\Visual Question Answering}

\author{Alberto Compagnoni$^{*1,2}$ \quad Marco Morini$^{*1}$ \quad Sara Sarto$^1$ \quad Federico Cocchi$^{1,2}$ \quad Davide Caffagni$^1$ \\ Marcella Cornia$^1$ \quad Lorenzo Baraldi$^1$ \quad Rita  Cucchiara$^{1}$  \\
$^1$University of Modena and Reggio Emilia, Italy \quad $^2$University of Pisa, Italy\\
{\tt\small $^1$\{name.surname\}@unimore.it \quad $^2$\{name.surname\}@phd.unipi.it}
\\
{\tt\small \href{https://aimagelab.github.io/ReAG/}{aimagelab.github.io/ReAG}}
}

\begin{document}
\maketitle
\begin{abstract}
Multimodal Large Language Models (MLLMs) have shown impressive capabilities in jointly understanding text, images, and videos, often evaluated via Visual Question Answering (VQA). However, even state-of-the-art MLLMs struggle with domain-specific or knowledge-intensive queries, where relevant information is underrepresented in pre-training data. Knowledge-based VQA (KB-VQA) addresses this by retrieving external documents to condition answer generation, but current retrieval-augmented approaches suffer from low precision, noisy passages, and limited reasoning. To address this, we propose \textbf{\ours}, a novel Reasoning-Augmented Multimodal RAG approach that combines coarse- and fine-grained retrieval with a critic model that filters irrelevant passages, ensuring high-quality additional context. The model follows a multi-stage training strategy leveraging reinforcement learning to enhance reasoning over retrieved content, while supervised fine-tuning serves only as a cold start. Extensive experiments on Encyclopedic-VQA and InfoSeek demonstrate that \ours significantly outperforms prior methods, improving answer accuracy and providing interpretable reasoning grounded in retrieved evidence.  % Our source code is publicly available at: {\small{\url{https://github.com/aimagelab/ReAG}}}.
\blfootnote{$^*$Equal contribution.}
\end{abstract}    
\section{Introduction}
\label{sec:intro}

Multimodal Large Language Models (MLLMs)~\cite{alayrac2022flamingo, liu2023visual, liu2023improved, bai2025qwen2} unify tasks involving multiple modalities, such as text, images and videos~\cite{caffagni2024r,sarto2025image}. Many of these tasks can be framed as Visual Question Answering (VQA)~\cite{goyal2017making,antol2015vqa,wu2017visual}, where a query may require understanding visual content, and the model must generate a faithful, correctly formatted response. Despite their broad pre-training, state-of-the-art MLLMs struggle with underrepresented, domain-specific queries~\cite{chen2023can, mensink2023encyclopedic}. This problem, known as Knowledge-based VQA (KB-VQA)~\cite{marino2019ok}, is commonly addressed by enriching MLLMs with domain-specific information from external sources, \ie, via Retrieval-Augmented Generation (RAG)~\cite{lewis2020retrieval}.

\begin{figure}[t]
    \centering
    \includegraphics[width=0.98\linewidth]{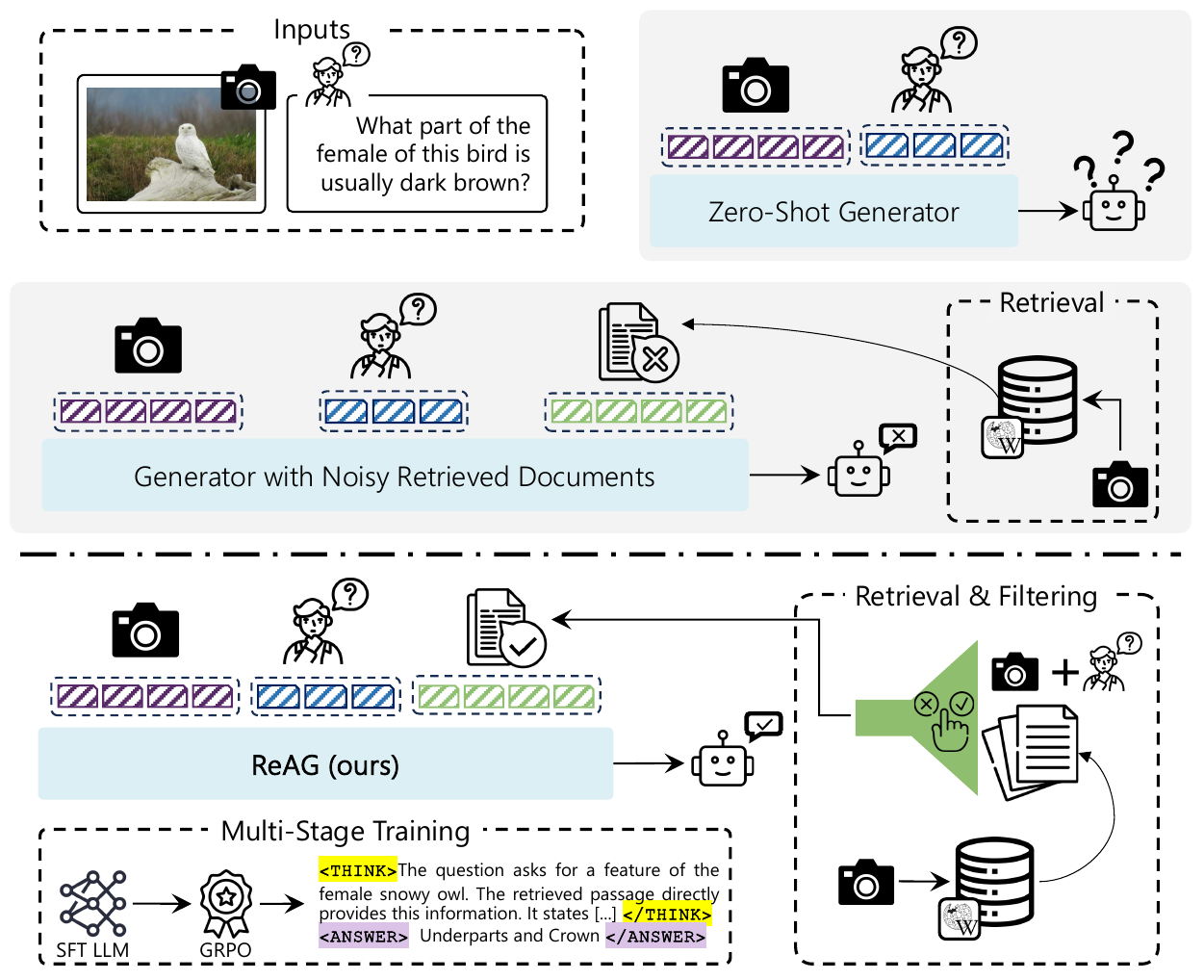}
    \vspace{-0.1cm}
    \caption{Comparison between Zero-Shot (ZS) MLLMs, retrieval-augmented models, and \ours. ZS MLLMs lack specialized knowledge and fail on domain-specific queries (top). Retrieval-augmented models introduce external context but often add noisy or irrelevant passages (middle). \ours overcomes this with a filtering stage over retrieved content and a multi-stage training strategy to enhance reasoning over passages.
    }
    \vspace{-0.3cm}
    \label{fig:first}
\end{figure}

Despite impressive results~\cite{yan2024echosight, wu2025towards, zhang2024mr, hong2025knowledge, cocchi2025augmenting, yuan2025mkg}, this setting still presents significant open challenges. One lies in information retrieval itself, as users’ queries can be extremely heterogeneous while the external knowledge-base can reach millions of documents in cardinality~\cite{chen2023can, mensink2023encyclopedic} -- thus lowering the recall of the retrieved results and adding noise to the MLLM input. This is further exacerbated by feature extraction and integration issues when queries and documents are multimodal, which is rapidly becoming usual~\cite{wei2023uniir, lin2024preflmr,caffagni2025recurrence}. Next, even assuming that the retrieved documents were relevant to the query, understanding them and extracting the right piece of information to generate the answer is not trivial~\cite{hua2025vision}.

To address these challenges, we propose \textbf{\ours}, short for \underline{\textbf{Re}}asoning-\underline{\textbf{A}}ugmented \underline{\textbf{G}}eneration, a novel multimodal retrieval-augmented generation approach that (i) mitigates low-recall and noisy retrieval by employing a multi-level retrieval pipeline followed by a critic model that effectively filters out irrelevant samples, and (ii) equips the MLLM with the capability of reasoning over retrieved results through a dedicated reinforcement learning training protocol. 

During retrieval, 
following common practice, \ours first employs an off-the-shelf multimodal encoder~\cite{radford2021learning, sun2024eva} for coarse-grained embedding-based retrieval, which achieves high recall when retaining many documents but suffers from low precision due to noise. To improve precision, a fine-grained retrieval stage focuses on the visual regions most relevant to the question. A critic model then classifies each passage as \textit{relevant} or \textit{irrelevant}, ensuring that only high-quality documents are passed to the generator (Fig.~\ref{fig:first}).

Building on advances in reasoning models~\cite{guo2025deepseek}, \ours further improves generation quality by allowing the model to produce explicit natural-language reasoning traces before the final answer.
Unlike prior multimodal RAG methods that rely on implicit reflection signals~\cite{asaiself, cocchi2025augmenting}, we train the model to create a full reasoning trace in natural language, without obeying any predefined pattern. To achieve this, \ours replaces the traditional supervised training in favor of a reinforcement learning framework inspired by GRPO~\cite{shao2024deepseekmath,yu2025dapo}, equipped with a reward scheme tailored for KB-VQA, to refine the model ability to reason over the user query and retrieved evidence, while using supervised fine-tuning only as a cold-start to establish initial reasoning behavior.

Experimentally, we evaluate the proposed approach on Encyclopedic-VQA~\cite{mensink2023encyclopedic} and InfoSeek~\cite{chen2023can}, two benchmarks containing question-answer pairs linked to Wikipedia-derived knowledge bases. Extensive experiments show that \ours substantially outperforms prior methods, not only improving answer accuracy but also generating explicit reasoning traces. These traces provide insight into the usefulness of retrieved passages and the steps leading to the final answer, offering full explainability of the model predictions.

\noindent In summary, the contributions of this work are as follows:
\begin{itemize}
    \item We propose \ours, a novel reasoning-augmented multimodal RAG model that combines coarse- and fine-grained retrieval with a critic model to improve precision and reduce noise injected to the generator. Notably, the critic is agnostic to the retrieval backbone, making it seamlessly applicable on top of any state-of-the-art retrieval engine.
   \item \ours trains the generator employing a multi-stage training strategy, leveraging SFT only as a cold start, followed by a reinforcement learning framework inspired by GRPO, with a reward scheme specifically designed for KB-VQA.
    \item We empirically validate \ours on two popular and challenging KB-VQA benchmarks, Encyclopedic-VQA and InfoSeek, where \ours reaches a new state-of-the-art.
\end{itemize}

\section{Related Work}
\label{sec:related}

\tinytit{Knowledge-based VQA}
The task requires models to answer questions that depend on external or specialized knowledge beyond the visual content of an image. Early datasets~\cite{marino2019ok,schwenk2022okvqa,shah2019kvqa}
targeted specialized reasoning. However, with the advent of more powerful MLLMs, these datasets have become insufficient for evaluating performance in realistic and knowledge-intensive settings.
To address this, benchmarks such as Encyclopedic-VQA~\cite{mensink2023encyclopedic} and InfoSeek~\cite{chen2023can} introduce more challenging, Wikipedia-scale scenarios requiring fine-grained and entity-specific 
reasoning over external knowledge, making retrieval essential.

\begin{figure*}[t]
    \centering    \includegraphics[width=0.98\linewidth]{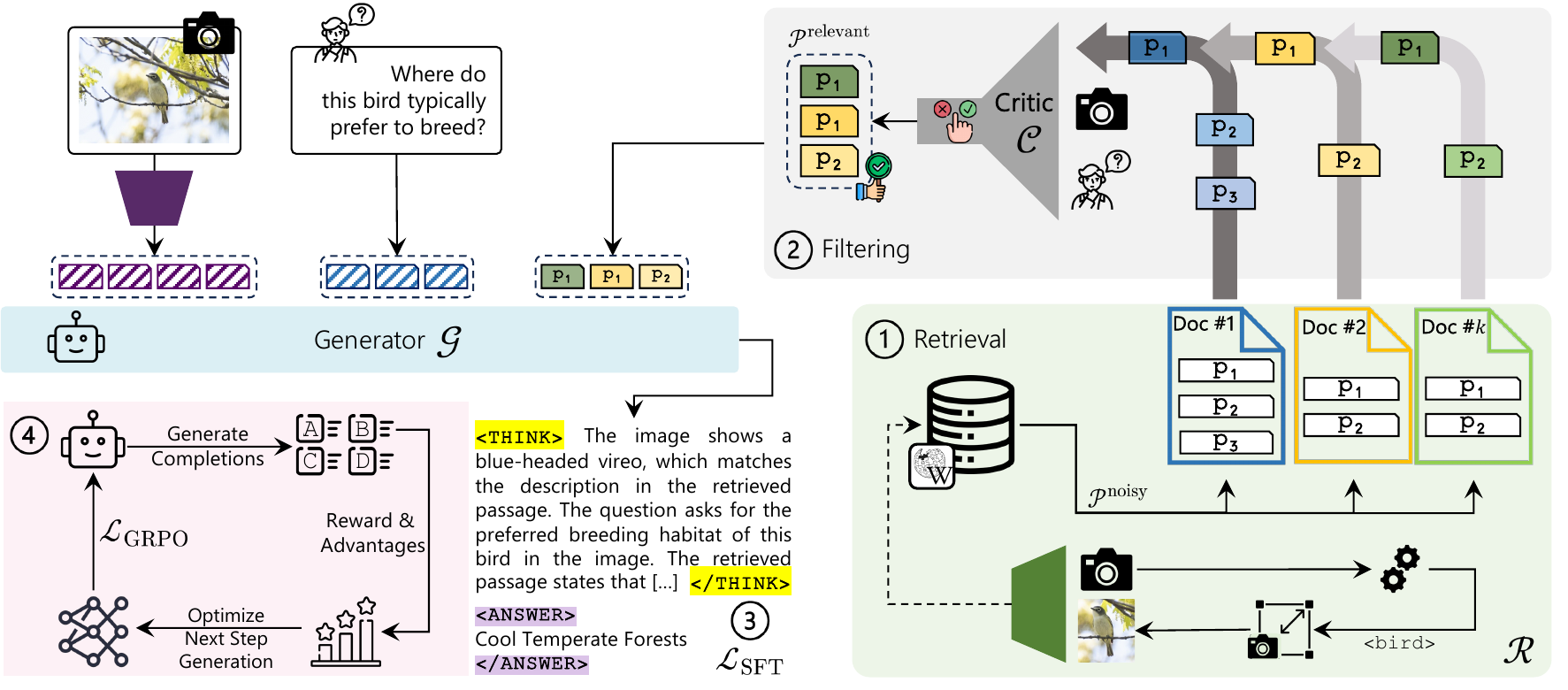}
    \vspace{-0.15cm}
    \caption{Overview of the proposed \ours model. A multi-level retriever module extracts noisy passages, which are refined by a critic model. The resulting relevant passages are fed to a generator trained via SFT and a reinforcement learning stage designed for the KB-VQA task.
    }
    \vspace{-0.3cm}
    \label{fig:model}
\end{figure*}
 
The RAG framework has become the standard approach for this task, retrieving relevant content from sources such as Wikipedia. 
One line of work focuses on enhancing the retrieval itself to obtain more accurate and less noisy results, as in WikiLLaVA~\cite{caffagni2024wiki}, which integrates external multimodal knowledge through a hierarchical retrieval framework. Other focuses on handling noisy retrieval, refining visual tokens~\cite{qi2024rora} or re-ranking retrieved textual passages before processing by the LLM~\cite{yan2024echosight}. 
A third direction strengthens model-level control: VLM-PRF~\cite{hong2025knowledge} employs external tools for knowledge filtering, while ReflectiVA~\cite{cocchi2025augmenting} uses control tokens to guide retrieval and knowledge assessment.

In this work, we first aim to reduce retrieval noise through effective filtering mechanisms and then empower the model with reasoning capabilities to critically evaluate retrieved knowledge before generating the final answer.

\tit{RL-based Strategies for LLMs and MLLMs}
As frontier LLMs advance, RL has emerged as a key paradigm for aligning model outputs with human values and desired behaviors~\cite{ouyang2022training,lee2023rlaif,rafailov2023direct}. Concurrently, the quality and diversity of training data remain crucial for robust alignment and generalization~\cite{zhou2023lima}.
Recently, GRPO~\cite{shao2024deepseekmath} has emerged as a promising approach to improve sample efficiency and training stability at scale. Several variants~\cite{yu2025dapo,liu2025understanding} revisit the underlying loss formulation to mitigate bias and enhance token-level optimization efficiency.

Inspired by the performance improvements achieved through GRPO-based methods, similar strategies have been extended to MLLMs~\cite{li2025self,wei2025unsupervised}. In particular, GRPO-CARE~\cite{chen2025grpo} enhances the coherence between intermediate reasoning traces and final outputs, leading to more reliable reasoning-grounded responses. Advanced LLM reasoning has also been explored in knowledge-intensive tasks. For example, Search-R1~\cite{jin2025search} integrates retrieval and reasoning for complex queries, while subsequent approaches extend this paradigm to multimodal search~\cite{jiang2024mmsearch,narayan2025deepmmsearch,wu2025mmsearch}.

In this work, we build upon these foundations by introducing a multi-stage reinforcement learning framework that operates on top of a supervised fine-tuned MLLM, enhancing its ability to reason effectively over retrieved evidence.

\section{Proposed Method}
\label{sec:method}

\tinytit{Task Definition}
In the standard VQA task, a multimodal LLM, referred to as the generator model $\mathcal{G}$, must answer a question $q$ about an image $\mathit{I_q}$. The task requires the model to understand the visual content and provide a correct answer. 
While MLLMs large-scale pretraining captures general knowledge, it may be insufficient for answering highly specific or domain-specific questions.

Knowledge-based VQA (KB-VQA) extends VQA by incorporating external knowledge. 
In our setting, the external knowledge base $\mathcal{KB}$ is a collection of $N$ multimodal documents (\eg,~Wikipedia pages) each containing textual passages and images. Formally, the knowledge base can be represented as
\begin{equation}
    \mathcal{KB} = \{d_1, \dots , d_N\}, \quad  d_i = (\mathcal{T}_i, I_i, \mathcal{P}_i),
\end{equation} 
where $\mathcal{T}_i$ is the metadata of the $i$-th document (\eg,~title and summary of a Wikipedia page), $I_i$ is the associated image, if present, and $\mathcal{P}_i$ are the textual passages of the document.

A retrieval model $\mathcal{R}$ is employed to select the top-$k$ relevant documents from $\mathcal{KB}$ and their associated passages $\mathcal{\tilde{P}} = \{ p_0, \dots, p_j \}$, which are then provided within the generator context window. 
Finally, the generator produces an answer $A$ conditioned on both the image, the question, and the retrieved passages, as follows:
\begin{equation}
\label{eq:answer}
A \sim \mathcal{G}(A \mid q, I_q,\{ p_0, \dots, p_j \}).
\end{equation}

During training, the generator $\mathcal{G}$ is optimized to maximize the likelihood of producing the correct answer given the visual and textual context. In particular, the retrieved passages $\tilde{\mathcal{P}}$ act as external conditioning signals that augment the understanding of the model of the visual scene and the question, enabling knowledge-grounded reasoning. The objective can thus be expressed as the negative log-likelihood of the ground-truth answer tokens, averaged over the training distribution, \ie
\begin{equation}
\small
\label{eq:gen loss}
\mathcal{L}(\theta)
= \mathbb{E}_{(I_q, q, \tilde{\mathcal{P}}) \sim \mathcal{D}}\left[
\frac{1}{|y|}\sum_{t=1}^{|y|} \log \mathcal{G}_\theta\!\big(y_t \mid q, I_q, \tilde{\mathcal{P}}, y_{<t}\big)
\right],
\end{equation} where $\mathcal{D}$ denotes the training distribution.

\tit{Methodology Summary} To address the existing challenges of retrieval-augmented models, \textbf{\ours} enhances KB-VQA performance by employing retrieval, filtering, and reasoning over the retrieved passages. The approach consists of two key components: a \textbf{critic model} that filters retrieval results, and a \textbf{generator} trained to reason over the filtered documents before producing the final answer. The overall pipeline is organized into four main stages: (1) a multi-level retrieval stage to gather candidate passages, (2) a filtering stage where the critic selects relevant content, (3) a cold-start supervised fine-tuning (SFT) stage to instill initial reasoning capabilities in the generator, and (4) a reinforcement learning stage to further refine reasoning and answer generation. Together, these components reduce noise and improve the ability of the generator to produce accurate, knowledge-intensive answers. An overview of our methodology is shown in Fig.~\ref{fig:model}.

\subsection{Retrieval Stage}
The retrieval stage identifies potentially informative passages related to the query image, which are subsequently filtered to provide the generator with relevant external knowledge for reasoning and answer generation. 
This process comprises two complementary steps: a \textit{coarse-grained} retrieval, which retrieves candidate documents based on the entire query image, and a \textit{fine-grained} retrieval, which performs retrieval using localized cues.
Notably, \ours is agnostic to the choice of retriever, so $\mathcal{R}$ can be any cross-modal encoder that maps the query image and either the metadata $\mathcal{T}_i$ or the image $I_i$\footnote{Depending on the test case, see Sec.~\ref{subsec:details} for details.} associated with each document $d_i$ into a shared embedding space. 
Relevance between queries and documents is then computed via cosine similarity.

\tit{Coarse-Grained Retrieval} 
An initial set of relevant textual passages, denoted as $\mathcal{P}^{\mathrm{cg}}$, is constructed by aggregating all the passages contained in the top-$k$ retrieved documents when using the original image $\mathit{I_q}$ as query to the retriever $\mathcal{R}$. Since each document contains a variable number of passages, the resulting collection is represented as $\mathcal{P}^{\mathrm{cg}} = \{p_1^{\mathrm{cg}}, \dots, p_m^{\mathrm{cg}}\}$, where $m$ denotes the total number of passages gathered from the top-$k$ documents.

\tit{Fine-Grained Retrieval}
To improve retrieval recall, we introduce a fine-grained retrieval stage that focuses on the specific visual region relevant to the question.
Given the input image $\mathit{I_q}$ and the question $q$, we identify a bounding box corresponding to the subject of the question, employing an off-the-shelf detection model. If such a region is detected, we crop the image accordingly, obtaining a focused image patch $\mathit{I_q}^{\mathrm{crop}}$.
This cropped image is then used as input to the retriever model $\mathcal{R}$, which computes relevance scores with respect to each document $d_i$, as in the coarse-grained stage. 

The top-$k$ documents retrieved in this stage form the fine-grained candidate passages, denoted as $\mathcal{P}^{\mathrm{fg}} = \{p_1^{\mathrm{fg}}, \dots, p_l^{\mathrm{fg}}\}$.
By restricting the visual input to the region of interest, this stage allows the retriever to focus on more fine-grained visual details, yielding passages that are more likely to be relevant to the specific question.

\tit{Final Set of Retrieved Passages} 
The documents comprising \(\mathcal{P}^{\mathrm{cg}}\) and \(\mathcal{P}^{\mathrm{fg}}\) are merged and ranked by their relevance scores, and all passages contained in the top-$k$ ranked documents are retained to form the final set \(\mathcal{P}^{\mathrm{noisy}}\) from \(\mathcal{KB}\).

\subsection{Filtering}
After the retrieval steps, we obtain a set $\mathcal{P}^{\mathrm{noisy}}$ of passages from the $k$ retrieved documents. While increasing $k$ generally improves recall by including more potentially relevant passages, this typically comes at the cost of a lower precision, as the probability of introducing noisy information rises as well. To mitigate this, we design a critic model $\mathcal{C}$ to filter out irrelevant passages, resulting in a refined set of relevant passages  $\mathcal{P}^{\mathrm{relevant}}$.

\tit{Critic Model}
Given a question $q$ and its corresponding image $I_q$, the critic model $\mathcal{C}$ predicts if each retrieved textual passage in $\mathcal{P}^{\mathrm{noisy}}$ is useful for answering the question\footnote{The exact prompt can be found in the supplementary material.}. 
In \ours, the critic model is implemented as an autoregressive MLLM fine-tuned with a next-token prediction objective optimized on an annotated dataset. Specifically, starting from a subset of samples drawn from the dataset employed in~\cite{cocchi2025augmenting}, we extract tuples $(I_{q}, q, p, y)$, where $p$ is a textual passage to be evaluated and $y \in \{\textit{Yes}, \textit{No}\}$ indicates whether the passage is relevant. The critic model is trained to predict $y$ conditioned on $(I_{q}, q, p)$, enabling it to robustly discriminate between relevant and irrelevant passages.

At inference time, only passages yielding a positive prediction with probability above a threshold are kept, yielding the final subset of relevant passages $\mathcal{P}^{\mathrm{relevant}}$, defined as: 
\begin{equation}
\small
\label{eq:critic2}
\mathcal{P}^{\mathrm{relevant}} = 
\{\, p \in \mathcal{P}^{\mathrm{noisy}} \mid \Pr(\textit{Yes} | \mathcal{C}, q, I_{q}, p) > \text{thresh} \,\}.
\end{equation}

The resulting set $\mathcal{P}^{\mathrm{relevant}}$ is fed to the generator $\mathcal{G}$, which leverages these passages to produce the final answer (Eq.~\ref{eq:answer}).

\subsection{Generator Cold Start}
\label{sec:cold_start}
Following the approach popularized by DeepSeek-R1~\cite{guo2025deepseek}, we train our generator $\mathcal{G}$ using a multi-stage strategy. The initial stage is designed to enhance the reasoning and zero-shot capabilities of the model, while mitigating potential instabilities during the subsequent reinforcement learning stage.
Unlike standard SFT, which focuses solely on answer prediction, our cold-start phase exposes $\mathcal{G}$ to explicit reasoning trajectories that link the visual content, retrieved passages, and the question.

\tit{Collecting Reasoning Traces}
To achieve this, we fine-tune $\mathcal{G}$ using high-quality reasoning data. Starting from the same subset used for training the critic model, we extend each tuple $(I_{q}, q, p, y)$ with a reasoning trace $tr$.
Specifically, each tuple is provided as input to an MLLM, which is prompted to generate an explicit reasoning trace that logically explains how the passage $p$ contributes to answering the question $q$ given the image $I_{q}$. To guide this reasoning, the prompt includes both the final answer and the relevance label $y$, indicating whether the reasoning should be grounded in the passage or not. By explicitly conditioning on these signals, the MLLM produces structured reasoning traces that reflect a coherent inference process from the evidence to the answer. These traces are used as supervision for the cold-start fine-tuning of the generator $\mathcal{G}$.

\tit{Training Protocol} 
Having collected the reasoning-augmented dataset, the generator $\mathcal{G}$ is trained to optimize both its reasoning ability and answer accuracy.
To guide the model towards structured reasoning behavior, we encourage a templated output format, where the reasoning trace and the final answer are delimited by special tokens which are explicitly added to the vocabulary, \ie
\begin{center}
  \texttt{<think>} \text{reasoning trace} \texttt{</think>} \\ 
  \texttt{<answer>} \text{answer} \texttt{</answer>}.  
\end{center}

This structure encourages the model to separate intermediate reasoning from the final prediction, improving interpretability and stability during generation.
Training is performed using a next-token prediction objective over both the reasoning trace and the final answer. The overall SFT loss balances the two components as follows:
\begin{equation}
    \mathcal{L}_\mathrm{SFT} = \alpha\mathcal{L}_{\mathrm{A}} + (1-\alpha)\mathcal{L}_{\mathrm{T}},
\end{equation}
where $\mathcal{L}_{\mathrm{A}}$ and $\mathcal{L}_{\mathrm{T}}$ denote the negative log-likelihood losses computed over the answer and reasoning trace, respectively.

\subsection{Generator RL Training} 
\label{sec:generator_training}
While supervised fine-tuning on cold-start data equips the generator with basic reasoning skills and coherent chain-of-thought generation, we further enhance its quality and robustness through a subsequent reinforcement learning stage.

\tit{Task-specific RL with Retrieved Passages} 
Our generator model is optimized with a custom objective inspired by GRPO~\cite{shao2024deepseekmath}, incorporating several modifications from DAPO~\cite{yu2025dapo}. Formally, the objective is defined as follows:
\begin{equation}
\small
\begin{aligned}
\mathcal{J}_{\text{GRPO}}(\theta) =\;& 
\mathbb{E}_{(I_q, q, p)\sim \mathcal{D}, \{o_i\}_{i=1}^N \sim \mathcal{G}_{\theta_\text{old}}(\cdot \mid I_q, q, p)} \Bigg[ \\
& \frac{1}{\sum_{i=1}^{N} |o_i|} 
  \sum_{i=1}^{N} \sum_{t=1}^{|o_i|} 
    \min \Big( 
        r_{i,t}(\theta) \hat{A}_{i,t}, \\
& \qquad \text{clip}\big( r_{i,t}(\theta), 1 - \varepsilon, 1 + \varepsilon \big) \hat{A}_{i,t} 
      \Big) 
\Bigg],
\end{aligned}
\label{eq:customgrpoloss}
\end{equation}
where $\mathcal{G}_{\theta_\text{old}}$ is the generator initialized from the SFT cold-start phase and $\mathcal{G}_\theta$ the generator after optional off-policy updates. Moreover, $\{o_i\}_{i=1}^N$ are the generated completions with  associated rewards $R_i$, and $\hat{A}_i$ the corresponding GRPO advantages~\cite{shao2024deepseekmath}.
In the formula, $r_{i,t}$ is computed as follows:
\begin{equation}
\small
    r_{i,t}(\theta)=\frac{\mathcal{G}_{\theta}(o_{i,t} \mid I_q, q, p ,o_{i,<t})}{\mathcal{G}_{\theta_{\text{old}}}(o_{i,t} \mid  I_q, q, p ,o_{i,<t})}.
\label{eq:off_policy_ratio}
\end{equation}

As in our setting the updates are never off-policy, $\mathcal{G}_{\theta}$ coincides with $\mathcal{G}_{\theta_{\text{old}}}$, thus the ratio $r_{i,t}(\theta)$ is always 1.
Unlike the GRPO formulation and following DAPO, we omit the KL divergence penalty, which overly constrains exploration of alternative reasoning trajectories. This also improves memory efficiency and training speed by removing the need for a reference model and an extra forward pass. Furthermore, the loss is computed at the token level, as averaging over variable-length sequences would reduce the contribution of tokens in longer sequences and weaken their updates.

At each training iteration, the generator $\mathcal{G}$ is prompted with $({I_q}, q, p)$ to autoregressively generate a group of $N$ completions $\{{o_i}\}_{i=1}^N$. Each generated completion is then evaluated by one or more rule-based reward functions, producing a reward $R_i$ to compute the advantage as:
\begin{equation}
\small
   \hat{A}_{i,t} = \frac{R_i - \text{mean}(\{R_i\}_{i=1}^N)}{\text{std}(\{R_i\}_{i=1}^N)}.
\label{eq:advantage_calculation}
\end{equation}

Advantages guide the policy updates: completions with above-average rewards have their likelihood increased, while those below the mean are down-weighted. This exposure allows the generator to explore diverse strategies for interacting with passages, gradually steering it toward producing reasoning trajectories that yield correct answers.

\tit{Rule-Based Reward Design} 
In our setting, we employ two complementary rule-based binary reward functions: a task-specific \textit{accuracy reward} and a \textit{format reward}. The task-specific accuracy reward $R_{\text{task}}(o_i)$ verifies whether a generated completion is correct by parsing the prediction according to the question type (\ie, numerical or textual, single- or multi-answer)\footnote{Further details are provided in the supplementary material.}.  The format reward $R_{\text{fmt}}(o_i)$ enforces adherence to the expected output template. Both functions return $1$ in case of success and $0$ otherwise.
The final reward associated to a completion $o_i$ is defined as a weighted sum of the two. Formally, 
\begin{equation}
\begin{aligned}
R_i = \gamma R_{\text{task}}(o_i) + \delta R_{\text{fmt}}(o_i),
\end{aligned}
\label{eq:total_reward}
\end{equation}
where $\gamma$ and $\delta$ are two hyperparameters.

\section{Experiments}
\begin{table*}[t]
\caption{VQA accuracy scores on the Encyclopedic-VQA test set and the InfoSeek validation set. The marker $\dagger$ represents our reproductions, while \colorbox{lightgray}{gray color} indicates models tested with non-comparable knowledge bases.}
  \vspace{-0.25cm}
  \centering
  \setlength{\tabcolsep}{.46em}
  \resizebox{\linewidth}{!}{
  \begin{tabular}{lc cc c cc c ccc}
   \toprule
    & & & & & \multicolumn{2}{c}{\textbf{E-VQA}} & & \multicolumn{3}{c}{\textbf{InfoSeek}} \\
    \cmidrule{6-7} \cmidrule{9-11}
     \textbf{Model} & \textbf{Generator} & & \textbf{Retriever} & & Single-Hop & All & & Unseen-Q & Unseen-E & All \\
    \midrule
    BLIP-2~\cite{li2023blip} & - & & - & & 12.6 & 12.4 & & 12.7 & 12.3 & 12.5 \\
    LLaVA-v1.5-7B~\cite{liu2023improved} & - & & - & & 16.3 & 16.9 & & 9.6 & 9.4 & 9.5 \\
    LLaVA-MORE-8B~\cite{cocchi2025llava} & - & & - & & 16.0 & 16.9 & & 8.3 & 8.9 & 7.8 \\
    Qwen2.5-VL-3B~\cite{bai2025qwen2} & - & & - & & 21.9 & 21.9 & & 18.9 & 17.7 & 18.3 \\
    Qwen2.5-VL-7B~\cite{bai2025qwen2} & - & & - & & 23.6 & 23.2 & & 22.8 & 24.1 & 23.7 \\
    \midrule
    \rowcolor{lightgray} 
    DPR$_\text{V+T}$~\cite{lerner2024cross} & Multi-passage BERT & & CLIP ViT-B/32 & & 29.1 & - & & - & - & 12.4 \\
    \rowcolor{lightgray} 
    RORA-VLM~\cite{qi2024rora} & LLaVA-v1.5-7B & & CLIP ViT-L/14 & & - & 20.3 & & 25.1 & 27.3 & - \\
    \rowcolor{lightgray} 
    EchoSight~\cite{yan2024echosight} & Mistral-7B/LLaMA-3-8B & & EVA-CLIP-8B & & 41.8 & - & & - & - & 31.3 \\
    \rowcolor{lightgray}
    CoMEM~\cite{wu2025towards} & Qwen2.5-VL-7B & & Custom VLM & & - & - & & 32.8 & 28.5 & - \\
    Wiki-LLaVA~\cite{caffagni2024wiki} & LLaVA-v1.5-7B & & CLIP ViT-L/14+Contriever & & 17.7 & 20.3 & & 30.1 & 27.8 & 28.9 \\
    EchoSight~\cite{yan2024echosight} & LLaMA-3.1-8B & & EVA-CLIP-8B & & 36.3 & 34.2 & & 30.0 & 30.7 & 30.4 \\
    ReflectiVA~\cite{cocchi2025augmenting} & LLaVA-MORE-8B & & EVA-CLIP-8B & & 35.5 & 35.5 & & 40.4 & 39.8 & 40.1 \\
    mR$^\text{2}$AG~\cite{zhang2024mr} & LLaVA-v1.5-7B & & CLIP ViT-L/14 & & - & - & & 40.6 & 39.8 & 40.2 \\
    VLM-PRF~\cite{hong2025knowledge} & LLaVA-MORE-8B & & EVA-CLIP-8B & & 36.3 & 35.5 & & 41.3 & 40.6 & 40.8 \\
    mKG-RAG~\cite{yuan2025mkg} & LLaVA-MORE-8B & & Custom VLM & & 38.4 & 36.3 & & 41.4 & 39.6 & 40.5 \\
    \midrule
    ReflectiVA~\cite{cocchi2025augmenting}\textsuperscript{\small$\dagger$} & Qwen2.5-VL-3B & & EVA-CLIP-8B & & 33.7 & 35.2 & & 39.6 & 38.1 & 38.9 \\
    VLM-PRF~\cite{hong2025knowledge} & Qwen2.5-VL-3B & & EVA-CLIP-8B & & 31.1 & 32.4 & & 39.7 & 38.8 & 39.0 \\
    \rowcolor{OurColor}
     &  & &  & & \textbf{41.3} & \textbf{42.9} & & \textbf{43.7} & \textbf{42.9} & \textbf{43.3} \\
    \rowcolor{OurColor}
   \multirow{-2}{*}{\textbf{\ours (Ours)}} & \multirow{-2}{*}{Qwen2.5-VL-3B} & & \multirow{-2}{*}{EVA-CLIP-8B} & & \inc{7.6} & \inc{7.7} & & \inc{4.0} & \inc{4.1} & \inc{4.3} \\
    \midrule
    ReflectiVA~\cite{cocchi2025augmenting}\textsuperscript{\small$\dagger$}  & Qwen2.5-VL-7B & & EVA-CLIP-8B & & 36.8 & 36.8 & & 43.5 & 44.3 & 43.9 \\
    
    VLM-PRF~\cite{hong2025knowledge} & Qwen2.5-VL-7B & & EVA-CLIP-8B & & 37.1 & 36.0 & & 43.3 & 42.7 & 42.8 \\
    VLM-PRF~\cite{hong2025knowledge} & InternVL3-8B & & EVA-CLIP-8B & & 40.1 & 39.2 & & 43.5 & 42.1 & 42.5 \\
    \rowcolor{OurColor}
     &  & &  & & \textbf{44.9} & \textbf{47.0} & & \textbf{48.3} & \textbf{46.2} & \textbf{47.2} \\ 
    \rowcolor{OurColor}
   \multirow{-2}{*}{\textbf{\ours (Ours)}} & \multirow{-2}{*}{Qwen2.5-VL-7B} & & \multirow{-2}{*}{EVA-CLIP-8B} & & \inc{4.8} & \inc{7.8} & & \inc{4.8} & \inc{1.9} & \inc{3.3} \\
  \bottomrule
  \end{tabular}
  }
\label{tab:results}
\vspace{-0.45cm}
\end{table*}

\subsection{Datasets and Evaluation Metrics}

\tinytit{Encyclopedic-VQA}
The Encyclopedic-VQA~\cite{mensink2023encyclopedic}
dataset contains 221k question-answer pairs, each linked to up to five images and covering 16.7k fine-grained entities.  
The questions are categorized into \textit{single-hop} and \textit{two-hop} types: single-hop questions can be answered using information from a single Wikipedia page, whereas two-hop questions require sequential retrieval across multiple pages. The dataset is divided into training, validation, and test splits comprising 1M, 13.6k, and 5.8k samples. All experiments are conducted on the test split, which includes 4.8k single-hop questions. The dataset contains an external knowledge base derived from Wikipedia, consisting of approximately 2M pages. Each comprises the article title, its textual sections, and associated images. In our experiments, we employ the original 2M-page knowledge base provided with the dataset.

\tit{InfoSeek} The InfoSeek dataset~\cite{chen2023can} consists of approximately 1.3M image-question-answer triplets corresponding to around 11k distinct Wikipedia pages. It is partitioned into training, validation, and test splits, containing roughly 934k, 73k, and 348k samples. Both validation and test sets include questions about unseen entities.
InfoSeek provides an external knowledge base of around 6M Wikipedia entities. Following previous works~\cite{caffagni2024wiki,cocchi2025augmenting,yan2024echosight}, experiments are conducted using a knowledge base of 100k pages.

\tit{Evaluation Metrics} 
We follow the original evaluation protocols provided with each dataset. For E-VQA, generated answers are evaluated using the BERT-based matching score (BEM)~\cite{bulian2022tomayto}, which measures semantic similarity of predicted and ground-truth answers. For InfoSeek, evaluation depends on the question type: we employ standard VQA accuracy~\cite{goyal2017making} as well as relaxed accuracy~\cite{methani2020plotqa}.

\subsection{Implementation Details}
\label{subsec:details}

\tinytit{Retrieval Details}
To retrieve potentially informative documents for a query image, we employ EVA-CLIP-8B~\cite{sun2024eva}. 
In the coarse-grained stage, the entire query image is encoded through EVA-CLIP to perform large-scale retrieval over the knowledge base.
For InfoSeek, we use an image-to-text retrieval setup that computes similarity between the query image and document metadata (\ie, the title of the page and the summary). For Encyclopedic-VQA, we adopt image-to-image retrieval, comparing the query image with the images inside Wikipedia pages.
In the fine-grained stage, we extract the main visual subject mentioned in the question using the spaCy library\footnote{\url{https://github.com/explosion/spaCy}} and localize it in the image via GroundingDINO~\cite{liu2023grounding}, whose bounding box is re-encoded through EVA-CLIP. Retrieval is done using the FAISS library~\cite{johnson2019billion}, with the top-$k$ results with $k=20$ retained at each stage.

\tit{Critic Model and Dataset}
Independently from the generator scale, our critic model builds upon Qwen2.5-VL-3B, fine-tuned on a curated subset of the ReflectiVA dataset~\cite{cocchi2025augmenting}\footnote{Further details on the training subset are given in the supplementary.}.
The model is trained for 1 epoch with a learning rate of $2 \times 10^{-6}$ and a global batch size of $32$.

\tit{Generator Training}
We build two versions of our generator, both based on Qwen2.5-VL~\cite{bai2025qwen2}, using the 3B and 7B model variants, and optimize them using the SFT plus RL training scheme. In the SFT phase, we use the same ReflectiVA~\cite{cocchi2025augmenting} subset employed to train the critic model, collecting reasoning traces from Qwen2.5-VL-7B. We set $\alpha=0.8$ to give more importance to final-answer tokens. The generator is trained for one epoch with SFT loss using AdamW~\cite{loshchilovdecoupled},
a learning rate of $2\times10^{-6}$ and an effective batch size of 128.
For RL post-training, we use the full Encyclopedic-VQA and InfoSeek sets from ReflectiVA, excluding samples from LLaVA-Instruct~\cite{liu2023improved}. Each batch includes 128 prompts with 8 completions per prompt. 
Training is conducted with Adam~\cite{kingma2015adam}, a learning rate $1\times10^{-6}$. Rewards weigh accuracy $\gamma=1.0$ over format $\delta=0.2$.
In all our experiments, we update the MLP adapter and LLM weights while keeping the vision encoder frozen. 

\subsection{Comparison with the State of the Art}
\begin{table}[t]
\caption{VQA accuracy scores on Encyclopedic-VQA and InfoSeek with OMGM as retrieval modality.}
\label{tab:results3}
  \vspace{-0.25cm}
  \centering
  \setlength{\tabcolsep}{.25em}
  \resizebox{\linewidth}{!}{
  \begin{tabular}{lc c cc c ccc}
   \toprule
    & & & \multicolumn{1}{c}{\textbf{E-VQA}}& & \multicolumn{3}{c}{\textbf{InfoSeek}} \\
    \cmidrule{4-4} \cmidrule{6-8}
     \textbf{Model} & \textbf{Generator} & & Single-Hop & & Un-Q & Un-E & All \\
    \midrule
    ReflectiVA~\cite{cocchi2025augmenting} & LLaVA-MORE-8B & & 41.8 & & 33.8 & 34.5 & 34.1 \\
    OMGM~\cite{yang2025omgm} & LLaVA-v1.5-7B & & 50.2 & & 43.5 & 43.5 & 43.5 \\
   \midrule
    ReflectiVA~\cite{cocchi2025augmenting} & Qwen2.5-VL-3B & & 44.6 & & 40.5 & 41.6 & 41.1 \\
      \rowcolor{OurColor}
    \textbf{\ours (Ours)} & Qwen2.5-VL-3B & & \textbf{49.1} & & \textbf{47.2} & \textbf{45.3} & \textbf{46.2} \\
    \midrule
    ReflectiVA~\cite{cocchi2025augmenting} & Qwen2.5-VL-7B & & 44.0 & & 43.3 & 44.0 & 43.6 \\
  
    \rowcolor{OurColor}
    \textbf{\ours (Ours)} & Qwen2.5-VL-7B & & \textbf{52.5} & & \textbf{50.3} & \textbf{48.2} & \textbf{49.2} \\
  \bottomrule
  \end{tabular}
  }
\vspace{-0.25cm}
\end{table}

\tinytit{Main Results} 
We present a comprehensive comparison of \ours on the E-VQA test set and the InfoSeek validation set against both zero-shot MLLMs and retrieval-augmented baselines. Specifically, we evaluate BLIP-2~\cite{li2023blip}, LLaVA-v1.5~\cite{liu2023improved}, LLaVA-MORE~\cite{cocchi2025llava}, and Qwen2.5-VL~\cite{bai2025qwen2} in a zero-shot setting, where the models receive only the query image and question as input. We further include retrieval-augmented approaches such as DPR~\cite{lerner2024cross}, RORA-VLM~\cite{qi2024rora}, EchoSight~\cite{yan2024echosight}, COMEM~\cite{wu2025towards}, WikiLLaVA~\cite{caffagni2024wiki}, mR$^\text{2}$AG~\cite{zhang2024mr}, mKG-RAG~\cite{yuan2025mkg}, ReflectiVA~\cite{cocchi2025augmenting}, and VLM-PRF~\cite{hong2025knowledge}. For fairness, we reproduce ReflectiVA using Qwen2.5-VL backbones at both 3B and 7B scales.

As shown in Table~\ref{tab:results}, zero-shot MLLMs, which rely solely on internal knowledge, are unable to accurately answer questions in knowledge-intensive benchmarks, underscoring the need for external retrieval. With the introduction of a retrieval pipeline, performance improves substantially. For example, on InfoSeek, the overall accuracy rises from around $20\%$ for the zero-shot Qwen2.5-VL-7B model to roughly $40\%$ with retrieval-augmented methods such as mKG-RAG. \ours further advances these results, achieving state-of-the-art performance on both E-VQA and InfoSeek across model scales. Specifically, on E-VQA \ours yields a $+7.7$ point gain over ReflectiVA when using Qwen2.5-VL-3B and a $+7.8$ point improvement over VLM-PRF when leveraging the stronger InternVL3-8B backbone.
Similar gains are observed on InfoSeek, with overall improvements of $+4.3$ and $+3.3$ points for Qwen2.5-VL-3B and 7B, respectively. These consistent improvements across both datasets demonstrate the effectiveness and robustness of our approach.

\begin{table}[t]
\caption{VQA accuracy scores on Encyclopedic-VQA and InfoSeek with oracle Wikipedia pages. }
  \label{tab:results_oracle}
    \vspace{-0.25cm}
  \centering
  \setlength{\tabcolsep}{.18em}
  \resizebox{\linewidth}{!}{
  \begin{tabular}{lcc c c ccc}
   \toprule
    & & & \multicolumn{1}{c}{\textbf{E-VQA}} & & \multicolumn{3}{c}{\textbf{InfoSeek}} \\
    \cmidrule{4-4} \cmidrule{6-8}
    \textbf{Model} & \textbf{Generator} & & Single-Hop & & Un-Q & Un-E & All \\
    \midrule
    Qwen2.5-VL-3B~\cite{bai2025qwen2} & Qwen2.5-3B & & 72.1 & & 47.0 & 43.0 & 44.9 \\
    Qwen2.5-VL-7B~\cite{bai2025qwen2} & Qwen2.5-7B & & 78.3 & & 41.6 & 41.3 & 41.4 \\
    \midrule
    ReflectiVA~\cite{cocchi2025augmenting} & Qwen2.5-VL-3B & & 72.9 & & 53.4 & \textbf{53.9} & 53.7 \\
    \rowcolor{OurColor}
    \textbf{\ours (Ours)} & Qwen2.5-VL-3B & & \textbf{79.0} & & \textbf{55.1} & 53.3 & \textbf{54.2} \\
    \midrule
    Wiki-LLaVA~\cite{caffagni2024wiki} & LLaVA-v1.5-7B & & 38.5 & & 52.7 & 50.3 & 51.5 \\
    
    ReflectiVA~\cite{cocchi2025augmenting} & LLaVA-MORE-8B & & 75.2 & & 57.8 & 57.4 & 57.6 \\
    ReflectiVA~\cite{cocchi2025augmenting} & Qwen2.5-VL-7B & & 71.3 & & 56.0 & 56.0 & 56.0 \\
    \rowcolor{OurColor}
    \textbf{\ours (Ours)} & Qwen2.5-VL-7B & & \textbf{81.5} & & \textbf{60.7} & \textbf{58.9} & \textbf{59.7} \\
  \bottomrule
  \end{tabular}
  }
\vspace{-0.35cm}
\end{table}

\begin{figure}
    \centering
    \includegraphics[width=0.99\linewidth]{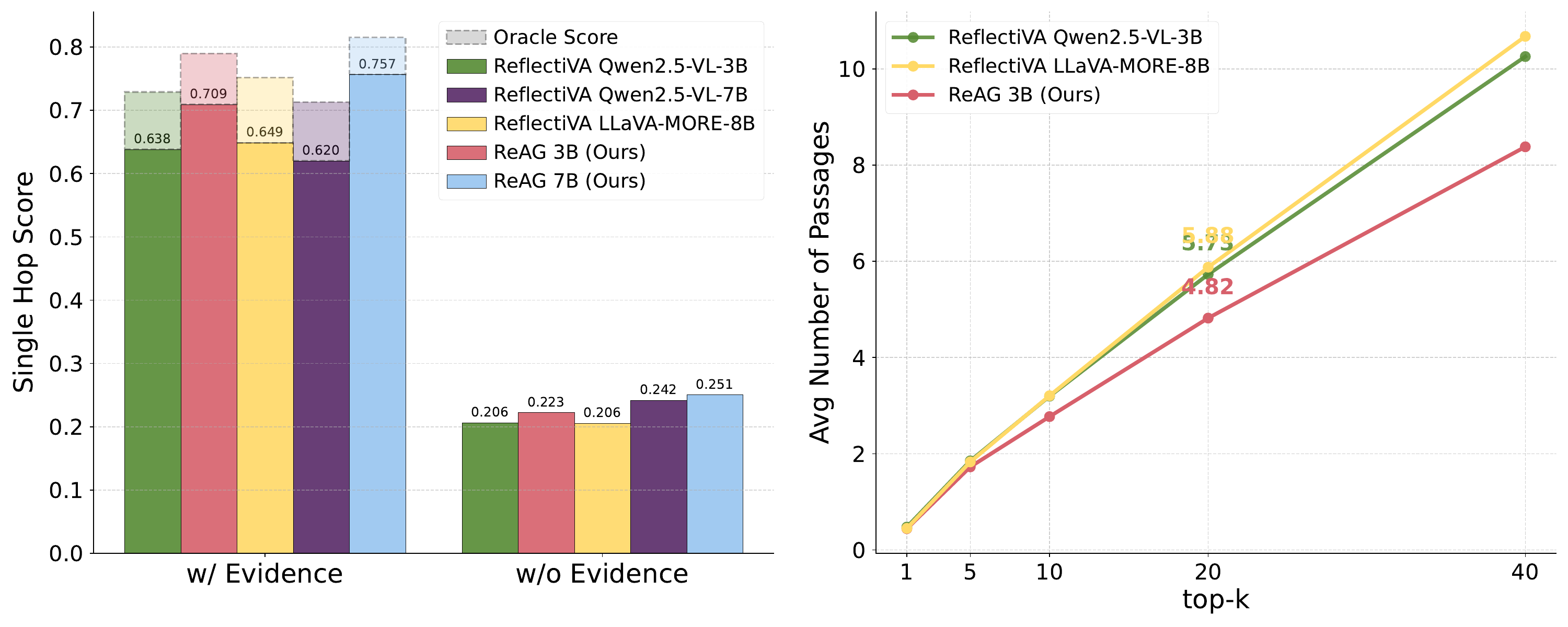}
    \vspace{-0.3cm}
    \caption{Comparison of performance on E-VQA with and without evidence, including oracle upper bounds (left). Analysis on average number of passages retained at different top‑$k$ values (right).}
    \label{fig:graphs}
    \vspace{-0.3cm}
\end{figure}

\begin{table*}[t]
\caption{Ablation study results on Encyclopedic-VQA and InfoSeek to validate the effectiveness of our model component.}
  \vspace{-0.2cm}
  \centering
  \setlength{\tabcolsep}{.25em}
  \resizebox{\linewidth}{!}{
  \begin{tabular}{lc cc ccc c cccc c c c ccc}
   \toprule
     & && \multicolumn{3}{c}{\textbf{Retrieval Pipeline}} & & \multicolumn{3}{c}{\textbf{Generation Pipeline}} & & \multicolumn{1}{c}{\textbf{E-VQA}} & & \multicolumn{3}{c}{\textbf{InfoSeek}} \\
     \cmidrule{4-6} \cmidrule{8-10} \cmidrule{12-12} \cmidrule{14-16}
      & \textbf{Generator} && Coarse-grained & Fine-grained & Critic & & SFT & Reasoning Traces & RL & & Single-Hop & & Unseen-Q & Unseen-E & All \\
     \midrule 
     &  && - & - & - & & - & - & - & & 21.9 & & 18.9 & 17.7 & 18.3 \\
    & && \cmark & - & - & & - & - & - & & 19.2 & & 10.2 & 10.0 & 10.1 \\
    & && \cmark & - & \cmark & & - & - & - & & 38.0 & & 27.9 & 26.1 & 27.0 \\
    & && \cmark & \cmark & \cmark & & - & - & - & & 40.2 & & 28.1 & 26.1 & 27.1 \\
    \midrule   
    & && \cmark & \cmark & \cmark & & \cmark & - & - & & 39.3 & & 37.9 & 37.1 & 37.5 \\
    & && \cmark & \cmark & \cmark & & \cmark & \cmark & - & & 38.1 & & 41.9 & 40.6 & 41.3 \\
    & && \cmark & \cmark & \cmark & & \cmark & - & \cmark & & 39.5 & & 39.8 & 39.4 & 39.6 \\
    \rowcolor{OurColor}
    \textbf{\ours (Ours)} & Qwen2.5-VL-3B && \cmark & \cmark & \cmark & & \cmark & \cmark & \cmark & & \textbf{41.3} & & \textbf{43.7} & \textbf{42.9} & \textbf{43.3} \\
    \midrule
    \midrule
    & && \cmark & \cmark & \cmark & & - & - & - & & 41.7 & & 29.3 & 27.8 & 28.5 \\
    & && \cmark & \cmark & \cmark & & \cmark & \cmark & - & & 42.0 & & 42.0 & 40.7 & 41.4 \\
    \rowcolor{OurColor}
     \textbf{\ours (Ours)}  & Qwen2.5-VL-7B && \cmark & \cmark & \cmark & & \cmark & \cmark & \cmark & & \textbf{44.9} & & \textbf{48.3} & \textbf{46.2} & \textbf{47.2} \\
  \bottomrule
  \end{tabular}
  }
\label{tab:ablation}
\vspace{-0.2cm}
\end{table*}

\begin{figure*}[t]
\begin{minipage}{0.325\linewidth}
\scriptsize{\textbf{Q}: What is the closest upper taxonomy of this bird?\vspace{0.05cm}}\\
\begin{minipage}{0.443\linewidth}
\includegraphics[width=1.\linewidth,height=0.88\linewidth]{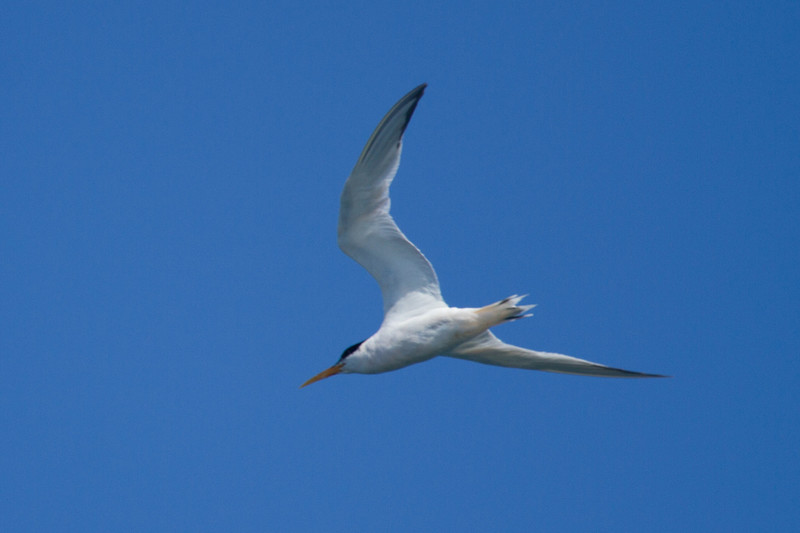}
\end{minipage}
\hfill
\begin{minipage}{0.53\linewidth}
\scriptsize{
\textbf{Qwen2.5-VL-7B (ZS)~\cite{bai2025qwen2}}:\\
The closest taxonomy of this bird is the family Laridae. \textcolor{red}{\xmark} \\
\textbf{ReflectiVA~\cite{cocchi2025augmenting}}:\\
Sterna \textcolor{red}{\xmark} \\
\textbf{\ours (Ours):}\\
Thalasseus\textcolor[HTML]{00b050}{\cmark}
}
\end{minipage}
\end{minipage}
\hspace{0.02cm}
\begin{minipage}{0.325\linewidth}
\scriptsize{\textbf{Q}:Which road, railway or canal does this bridge carry?\vspace{0.05cm}}\\
\begin{minipage}{0.443\linewidth}
\includegraphics[width=1.\linewidth,height=0.88\linewidth]{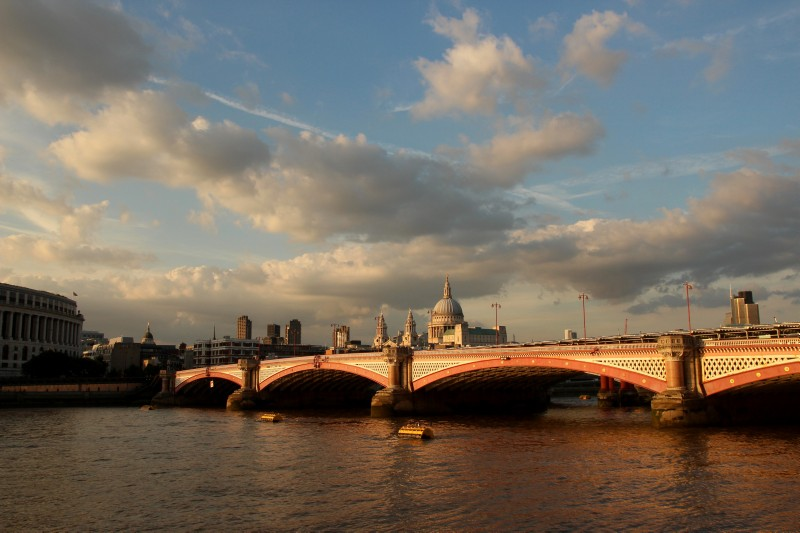}
\end{minipage}
\hfill
\begin{minipage}{0.53\linewidth}
\scriptsize{
\textbf{Qwen2.5-VL-7B (ZS)~\cite{bai2025qwen2}}:\\
The bridge in the picture is Blackfriars Bridge. \textcolor{red}{\xmark} \\
\textbf{ReflectiVA~\cite{cocchi2025augmenting}}: \\ 
Road \textcolor{red}{\xmark} \\
\textbf{\ours (Ours):}\\
A201 \textcolor[HTML]{00b050}{\cmark}
}
\end{minipage}
\end{minipage}
\hspace{0.02cm}
\begin{minipage}{0.325\linewidth}
\scriptsize{\textbf{Q:} What is the brand of this vehicle? \vspace{0.05cm}}\\
\begin{minipage}{0.443\linewidth}
\includegraphics[width=1.\linewidth,height=0.88\linewidth]{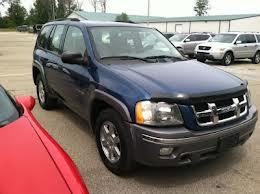}
\end{minipage}
\hfill
\begin{minipage}{0.53\linewidth}
\scriptsize{
\textbf{Qwen2.5-VL-7B (ZS)~\cite{bai2025qwen2}}:\\
The vehicle in the picture is a Ford. This can be determined [..] \textcolor{red}{\xmark} \\
\textbf{ReflectiVA~\cite{cocchi2025augmenting}}:\\
Ford \textcolor{red}{\xmark} \\
\textbf{\ours (Ours):}\\
Isuzu \textcolor[HTML]{00b050}{\cmark}
}
\end{minipage}
\end{minipage}
\vspace{-0.25cm}
\caption{Qualitative results on InfoSeek image-question pairs comparing \ours, ReflectiVA~\cite{cocchi2025augmenting}, and the corresponding zero-shot model.}
\label{fig:qualitatives}
\vspace{-0.35cm}
\end{figure*}

\tit{Results with OMGM Retrieval Mode}
We also compare against the OMGM framework~\cite{yang2025omgm}, which adopts a coarse-to-fine, multi-stage retrieval strategy, leveraging an image-to-text summary retriever in the first step.
To ensure a fair comparison, in Table~\ref{tab:results3} we evaluate our method using the same retrieval modality. Notably, our method consistently outperforms OMGM~\cite{yang2025omgm} across both E-VQA and InfoSeek benchmarks.
With the Qwen2.5-VL-3B generator, our approach achieves substantial improvements over ReflectiVA~\cite{cocchi2025augmenting}, yielding gains of $+4.5$ and $+5.1$ points on E-VQA and InfoSeek. 
When scaling to Qwen2.5-VL-7B, performance further increases, reaching $52.5$ on E-VQA and $49.2$ on InfoSeek, surpassing OMGM by $2.3$ and $5.7$ points, respectively. 
These results indicate that, even when using only the initial retrieval step of OMGM, our critic-based filtering, together with \ours reasoning capabilities, leads to consistently higher performance compared to both prior methods and the full multi-stage retrieval of OMGM.

\tit{Results with Oracle Documents} 
We also experiment under an oracle setting (Table~\ref{tab:results_oracle}), where the ground-truth document (\ie, the Wikipedia page corresponding to the query) is provided. 
We compare results from zero-shot models (Qwen2.5-VL~\cite{bai2025qwen2} in both 3B and 7B variants), which take the entire Wikipedia pages as input, and retrieval-based methods (WikiLLaVA~\cite{caffagni2024wiki}, ReflectiVA~\cite{cocchi2025augmenting}, and \ours), which process retrieved passages through an additional model-specific filtering stage. In this configuration, \ours receives all passages from the oracle document, which are then passed to the critic model for filtering before being fed to the generator.

Notably, \ours achieves the best performance across both E-VQA and InfoSeek at all model scales. On the 3B variant, \ours outperforms ReflectiVA by $+6.1$ points on E-VQA, while on Infoseek the 7B version consistently improves over ReflectiVA by $+3.7$ and still surpasses it by $+2.1$ even when ReflectiVA employs a larger generator (LLaVA-MORE-8B).

\tit{Retrieval and Generation Pipeline Analysis} 
The performance of RAG-based approaches strongly depends on the presence of the evidence passage in the retrieved set, and on the number of passages provided to the generator. 

In Fig.~\ref{fig:graphs} (left), we evaluate the ability to produce the correct answer when the evidence passage is either present or absent in the context. \ours consistently outperforms all competitors of comparable scale, demonstrating that our reasoning-enhanced approach is robust even in the absence of direct evidence. Each model is also accompanied by its oracle performance, clearly showing that \ours consistently gets closer to the oracle upper bound than other approaches.

In Fig.~\ref{fig:graphs} (right), we instead report the average number of passages passed to the generation at different $k$ values, comparing the filtering behavior of ReflectiVA~\cite{cocchi2025augmenting} with that of our critic model. \ours reduces noise introduced in the generator context by achieving a reduction of $18.0\%$ and $15.9\%$ in the number of passages compared to ReflectiVA based on LLaVA-MORE-8B and Qwen2.5VL-3B respectively, further emphasizing the advantage of our filtering pipeline.

\tit{Qualitative Results} 
In Fig.~\ref{fig:qualitatives}, we present a qualitative comparison on image-question pairs from InfoSeek. Notably, the zero-shot model tends to produce longer and detailed answers, whereas ReflectiVA and \ours generate responses that follow the dataset-specific format. Overall, the results consistently demonstrate that \ours answers accurate and outperforms competing approaches.

\subsection{Ablation Studies} 
We finally perform an ablation study by progressively enabling key components of the final architecture. Table~\ref{tab:ablation} reports results at each stage with $k$ equal to 20, evaluating the effect of the critic and fine-grained retrieval modules, as well as design choices in the generation pipeline.

We first examine the zero-shot setup under different retrieval configurations, using the 3B-scale model. As shown in the first two rows of Table~\ref{tab:ablation}, directly passing all passages from the top-20 documents into the generator severely degrades performance due to excessive noise. Introducing the critic model (third row) effectively filters irrelevant passages, yielding more than a twofold gain. Adding the fine-grained retriever (fourth row) further improves retrieval recall and yields additional gains on both datasets.

Fixing the retrieval pipeline to its final configuration, we then assess the impact of different generation strategies. As shown in~\cite{guo2025deepseek}, introducing a cold-start phase can help prepare the model for reasoning by exposing it to intermediate traces before full supervision. Results show that applying reinforcement learning after this cold-start phase outperforms standard SFT, indicating that the cold-start phase effectively prepares the model for multi-step reasoning, allowing the RL algorithm to operate on a model already prepared for structured thinking. 
A similar trend is observed with the 7B variant, where both training phases contribute significantly to the final performance, further validating the robustness of the proposed \ours pipeline.

\section{Conclusion}
We have presented \ours, a multimodal retrieval-augmented approach designed for the KB-VQA task. Our method demonstrates that incorporating reasoning and filtering retrieved passages significantly improves answer quality by reducing the noise introduced by irrelevant passages and producing a structured reasoning process when generating the final answer. Extensive experimental results on both Encyclopedic-VQA and Infoseek show that \ours consistently outperforms existing methods, confirming the effectiveness of our retrieval and reasoning strategies.

\section*{Acknowledgments}
We acknowledge the CINECA award under the ISCRA initiative, for the availability of high-performance computing resources. This work has been supported by the EU Horizon project ``ELLIOT -  European Large Open Multi-Modal Foundation Models For Robust Generalization On Arbitrary Data Streams'' (No. 101214398), by the EuroHPC JU project ``MINERVA'' (GA No. 101182737), and by the PRIN 2022-PNRR project ``MUCES'' (CUP E53D23016290001) funded by the EU - NextGenerationEU.

{
    \small
    \bibliographystyle{ieeenat_fullname}
    \bibliography{bibliography}
}

% \begin{table}
%   \caption{Results.   Ours is better.}
%   \label{tab:example}
%   \centering
%   \begin{tabular}{@{}lc@{}}
%     \toprule
%     Method & Frobnability \\
%     \midrule
%     Theirs & Frumpy \\
%     Yours & Frobbly \\
%     Ours & Makes one's heart Frob\\
%     \bottomrule
%   \end{tabular}
% \end{table}

% \begin{figure}[t]
%   \centering
%   \fbox{\rule{0pt}{2in} \rule{0.9\linewidth}{0pt}}
%    %\includegraphics[width=0.8\linewidth]{egfigure.eps}
% 
%    \caption{Example of caption.
%    It is set in Roman so that mathematics (always set in Roman: $B \sin A = A \sin B$) may be included without an ugly clash.}
%    \label{fig:onecol}
% \end{figure}

% WARNING: do not forget to delete the supplementary pages from your submission 
\clearpage

\maketitlesupplementary
\appendix

\section{Additional Implementation Details}
\label{supp:details}

This section contains detailed descriptions of our implementation, including model architectures details, training setup, and evaluation protocols.

\tit{Retrieval Details}
As mentioned in the main paper, to retrieve candidate documents for each query image, we adopt a two-stage retrieval pipeline based on EVA-CLIP-8B. In the coarse-grained stage, the full query image is encoded and matched against the knowledge base. Instead, in the fine-grained retrieval phase, the subject of the question, corresponding to the visual entity in the image, is extracted using SpaCy\footnote{In particular, we use the \texttt{en\_core\_web\_sm} model available at \url{https://spacy.io/models/en}.}. According to the structure of query questions, extraction prioritizes noun phrases starting with demonstratives like ``this'' or ``these'', followed by nouns serving as objects of prepositions (\eg, ``of'', ``by'', ``in'', ``from'', etc.). If neither pattern is found, the last noun in the question is used as a fallback. This approach ensures that the most relevant entity is reliably identified for retrieval and reasoning.

\tit{Critic Model and Dataset}
The critic model builds upon Qwen2.5-VL-3B and is fine-tuned on a curated subset of the ReflectiVA dataset~\cite{cocchi2025augmenting}. Specifically, we select 1M samples, balanced between InfoSeek and Encyclopedic-VQA, where each sample is paired with a passage labeled as relevant or irrelevant. To encourage more robust discrimination, irrelevant passages are further divided into \textit{soft negatives} (\ie, semantically related but unhelpful passages) and \textit{hard negatives} (\ie, irrelevant passages that exhibit high similarity to the relevant ones), in proportions of 30\% and 70\% respectively. During training, we fine-tune only the visual projector and the LLM, while keeping the vision encoder frozen.

At inference time, given $(I_{q}, q, p)$ (where $I_{q}$ and $q$ are the input image and question, and $p$ is the passage to evaluate), we apply a \emph{yes}-probability threshold equal to 0.1 (cf. Eq.~\ref{eq:critic2} of the main paper). This conservative threshold ensures that the critic $\mathcal{C}$ filters out only those passages for which it is highly confident they are irrelevant.

\tit{Generator Training}
Our generator is based on Qwen2.5-VL~\cite{bai2025qwen2}, employing the 3B and 7B scale. As mentioned in the main paper, we employ a multi-stage training strategy: first an SFT stage that serves as cold start for a subsequent RL-based training stage.  Specifically, for both the cold-start and RL stages, we apply a cosine learning rate schedule with 450 and 150 warm-up steps, respectively. Weight decay is set to 0.01 during cold-start and removed during RL. 
SFT fine-tuning of the 3B and 7B models requires roughly 24 hours on 16 and 64 NVIDIA A100 64GB GPUs, respectively.
In the RL stage, completions are generated with a temperature equal to $1.0$ and a repetition penalty of 1.05, using vLLM~\cite{kwon2023efficient}. 

During training with the custom GRPO loss (cf. Eq.~\ref{eq:customgrpoloss} of the main paper), $\mathcal{G}_\theta$ and ${\mathcal{G}_\theta}_{\text{old}}$ share the same weights, although they operate as separate models. ${\mathcal{G}_\theta}_{\text{old}}$ runs inside the vLLM worker and remains frozen, and gradients are applied only to $\mathcal{G}_\theta$. After each backward pass, we synchronize ${\mathcal{G}_\theta}_{\text{old}}$ with the updated weights of $\mathcal{G}_\theta$. Fine-tuning the 3B and 7B models with our RL strategy takes roughly 48 hours on 32 and 64 NVIDIA A100 64GB GPUs, respectively. We select the best checkpoint based on the best task-specific accuracy on a held-out validation split. All runs employ DeepSpeed ZeRO-3~\cite{rajbhandari2020zero} and gradient checkpointing. Our training codebase builds on Open-R1~\cite{openr1} and TRL~\cite{vonwerra2022trl}.  

\section{Reward Design}
As discussed in Sec.~\ref{sec:generator_training} of the main paper, we employ a task-specific accuracy reward. The reward function evaluates only the final answer rather than the intermediate reasoning. To extract the predicted answer, we first search for content enclosed within the \texttt{<answer></answer>} tags. If no such content is found, we extract all text following the first \texttt{<answer>} tag. If this is unsuccessful, we instead use the text following the \texttt{</think>} tag. When none of these patterns appear, the entire model output is used. 

In every case, format-specific special tokens are removed. We then apply the same normalization procedure used in the InfoSeek and Encyclopedic-VQA evaluations, including the removal of articles, punctuation, extra whitespace, and capitalization, along with standardization of digits and contractions.
The final task-specific reward depends on the source dataset and task type, and is computed as follows:
\begin{equation*}
\small
R_{\text{task}}\big(\tilde{o}_i, {o}_i^*, \tau_i\big) =
\begin{cases}
\mathbbm{1}\big[\Psi_{\text{num}}(\tilde{o}_i,{o}_i^*)\big], & \text{if } \tau_i=\text{numerical},\\[1.5ex]
\mathbbm{1}\big[\text{IoU}(\tilde{o}_i,{o}_i^*)\ge 0.5\big], & \text{if } \tau_i=\text{multi},\\[1.5ex]
\mathbbm{1}\big[\tilde{o}_i = {o}_i^*\big], & \text{otherwise}.
\end{cases}
\end{equation*}
where $\tilde{o}_i$, ${o}_i^*$ and $\tau_i$ denote respectively the extracted prediction, the ground-truth answer and the task type of the $i$-th sample, and $\Psi_{\text{num}}$ evaluates success or failure in numerical match. When multiple alternative ground-truths are provided for a sample, we compute the reward with respect to each and take the maximum.
For samples from InfoSeek, we use exact string matching for \textit{entity} and \textit{time} questions, while \textit{numerical} questions are evaluated with $\Psi_{\text{num}}$:
\begin{equation*}
\small
\Psi_{\text{num}}(\tilde{o},{o}^*) =
\begin{cases}
|\tilde{o}-{o}^*|\le 0.1, & \text{if } \texttt{is\_scalar}(\tilde{o}) \\
& \quad \wedge \text{ }\texttt{is\_scalar}({o}^*),\\[1mm]
\tilde{o} \in {o}^*, & \text{if } \texttt{is\_scalar}(\tilde{o}) \\
& \quad \wedge \text{ }\texttt{is\_interval}({o}^*),\\[1mm]
\mathrm{IoU}(\tilde{o},{o}^*)\ge 0.5, & \text{if } \texttt{is\_interval}(\tilde{o}) \\
& \quad \wedge \text{ }\texttt{is\_interval}({o}^*).
\end{cases}
\end{equation*}

For samples from Encyclopedic-VQA dataset, we adopt exact match scoring for \textit{single-answer} questions, while for \textit{multi-answer} questions the prediction is rewarded as correct only if intersection-over-union between predicted and ground-truth items reaches or surpasses 0.5. 

\begin{figure}[t]
\vspace{-0.1cm}
    \centering
    \includegraphics[width=0.99\linewidth]{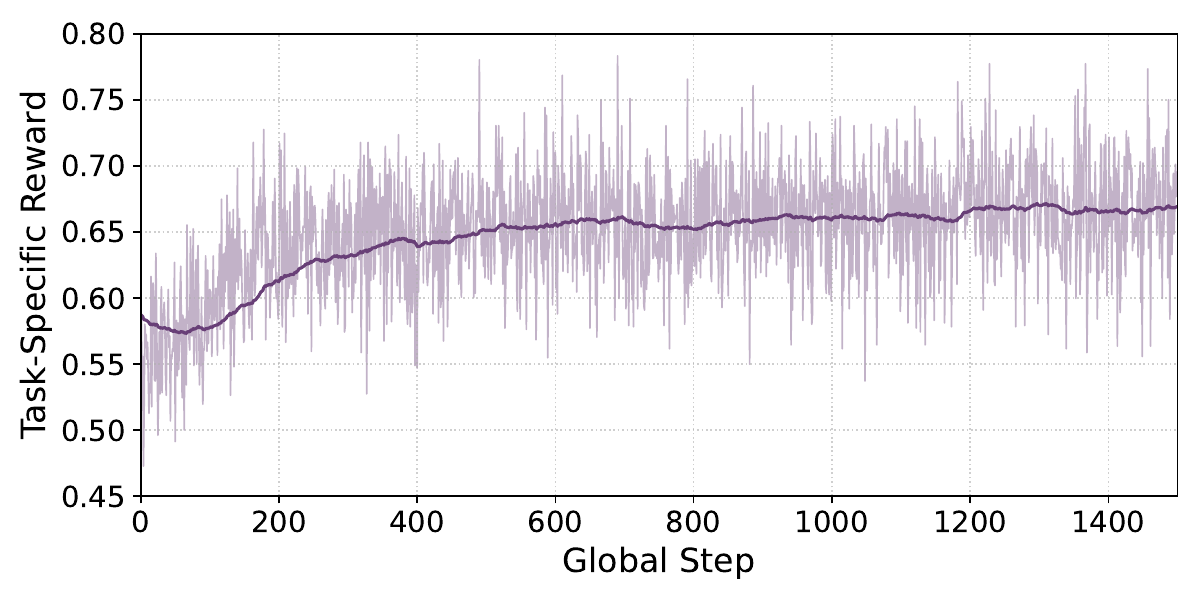}
    \vspace{-0.4cm}
    \caption{Task‑specific accuracy reward progression across training iterations of the \ours 7B generator.}
    \label{fig:reward_plot}
    \vspace{-0.3cm}
\end{figure}

The evolution of task-specific accuracy reward during training is reported in Fig.~\ref{fig:reward_plot}.

\section{Additional Experimental Results}
In this section, we provide additional experiments and analyses that complement the results reported in the main paper.

\subsection{Results with Google Lens Retriever} 
We further extend the analysis in the main paper by evaluating all methods under an alternative retrieval setup. Specifically, in Table~\ref{tab:results_lens}, we employ the Wikipedia pages retrieved by Google Lens~\footnote{A visual recognition service by Google, available at \url{https://lens.google.com/}.} when provided with the query image for each question of Encyclopedic-VQA, which have been officially released along with the dataset. Even though \ours does not use the fine-grained retriever in this setting, it consistently outperforms ReflectiVA~\cite{cocchi2025augmenting} across different generator scales. Notably, \ours at the 3B scale performs comparable to ReflectiVA at the 8B scale and HAMMR~\cite{castrejon2024hammr} at 55B. In addition, the stronger reasoning capabilities of \ours allow it to benefit substantially from improved retrieval quality, improving single-hop accuracy from 48.0 (3B) to 55.5 (7B), showing a notable gain of +7.5 points.

\begin{table}[t]
\caption{VQA accuracy scores on the Encyclopedic-VQA test set with Google Lens employed as retriever.}
\label{tab:results_lens}
  \vspace{-0.2cm}
  \centering
  \setlength{\tabcolsep}{.5em}
  \resizebox{0.97\linewidth}{!}{
  \begin{tabular}{lc c cc}
   \toprule
    & & & \multicolumn{2}{c}{\textbf{E-VQA}} \\
    \cmidrule{4-5}
     \textbf{Model} & \textbf{Generator} & & Single-Hop & All \\
    \midrule
    Qwen2.5-VL-3B~\cite{bai2025qwen2} & - & & 35.2 & - \\
    Qwen2.5-VL-7B~\cite{bai2025qwen2} & - & & 44.4 & - \\
    \midrule
    HAMMR~\cite{castrejon2024hammr} & PaLI-X-55B & & 47.8 & - \\
    mR$^\text{2}$AG~\cite{zhang2024mr} & LLaVA-v1.5-7B & & 55.9 & - \\
    ReflectiVA~\cite{cocchi2025augmenting} & LLaVA-MORE-8B & & 48.2 & 46.1 \\
   \midrule
    ReflectiVA~\cite{cocchi2025augmenting} & Qwen2.5-VL-3B & & 45.8 & 43.2 \\
     \rowcolor{OurColor}
    \textbf{\ours (Ours)} & Qwen2.5-VL-3B & & \textbf{48.0} & \textbf{48.0} \\
    \midrule
    ReflectiVA~\cite{cocchi2025augmenting} & Qwen2.5-VL-7B & & 48.3 & 46.8 \\
   
    \rowcolor{OurColor}
    \textbf{\ours (Ours)} & Qwen2.5-VL-7B & & \textbf{55.5} & \textbf{56.9} \\   
  \bottomrule
  \end{tabular}
  }
\vspace{-0.35cm}
\end{table}

\subsection{Varying the Number of Retrieved Documents}
\begin{figure}
   \centering
   \includegraphics[width=0.94\linewidth]{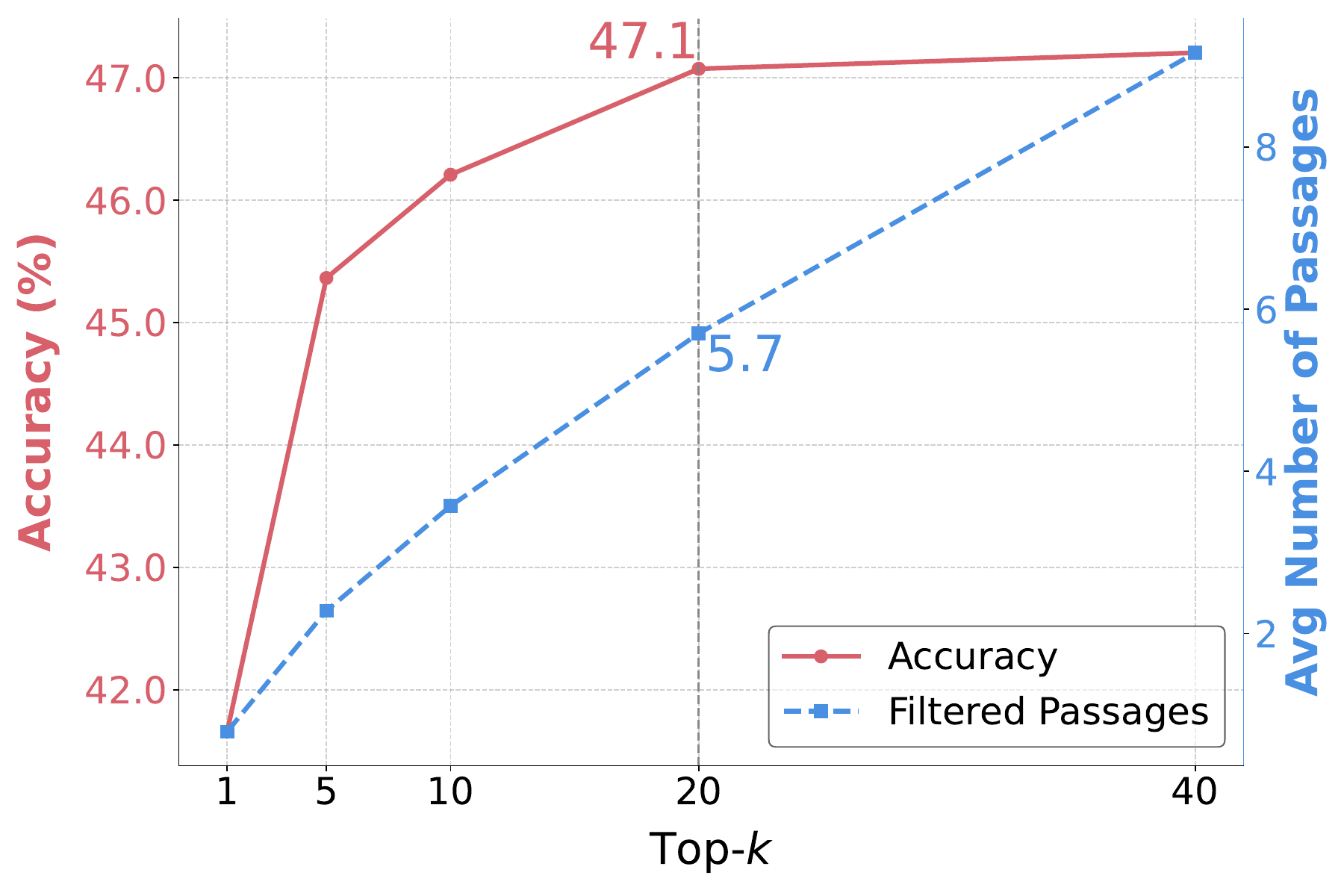}
   \vspace{-0.3cm}
    \caption{Performance of \ours 7B (red) and average number of filtered passages (blue) when varying the number $k$ of retrieved documents. Accuracy and number of passages are computed as the average of E-VQA and InfoSeek scores.}   \label{fig:k_analysys_avg}
    \vspace{-0.3cm}
\end{figure}

In Fig.~\ref{fig:k_analysys_avg}, we analyze the effect of varying the number of retrieved documents $k$ on the overall performance and on the average number of filtered passages fed to the generator. As shown, the model achieves the best results around $k=20$, which represents the optimal trade-off between coverage and noise, and is therefore adopted as the default retrieval depth in our pipeline. Retrieving too few documents results in insufficient contextual evidence, causing a drop in recall and limiting the ability of the model to access the necessary information. Conversely, increasing $k$ beyond this point does not yield meaningful performance gains while substantially inflating the computational cost of the filtering stage. 

\begin{figure}
    \centering
    \includegraphics[width=0.96\linewidth]{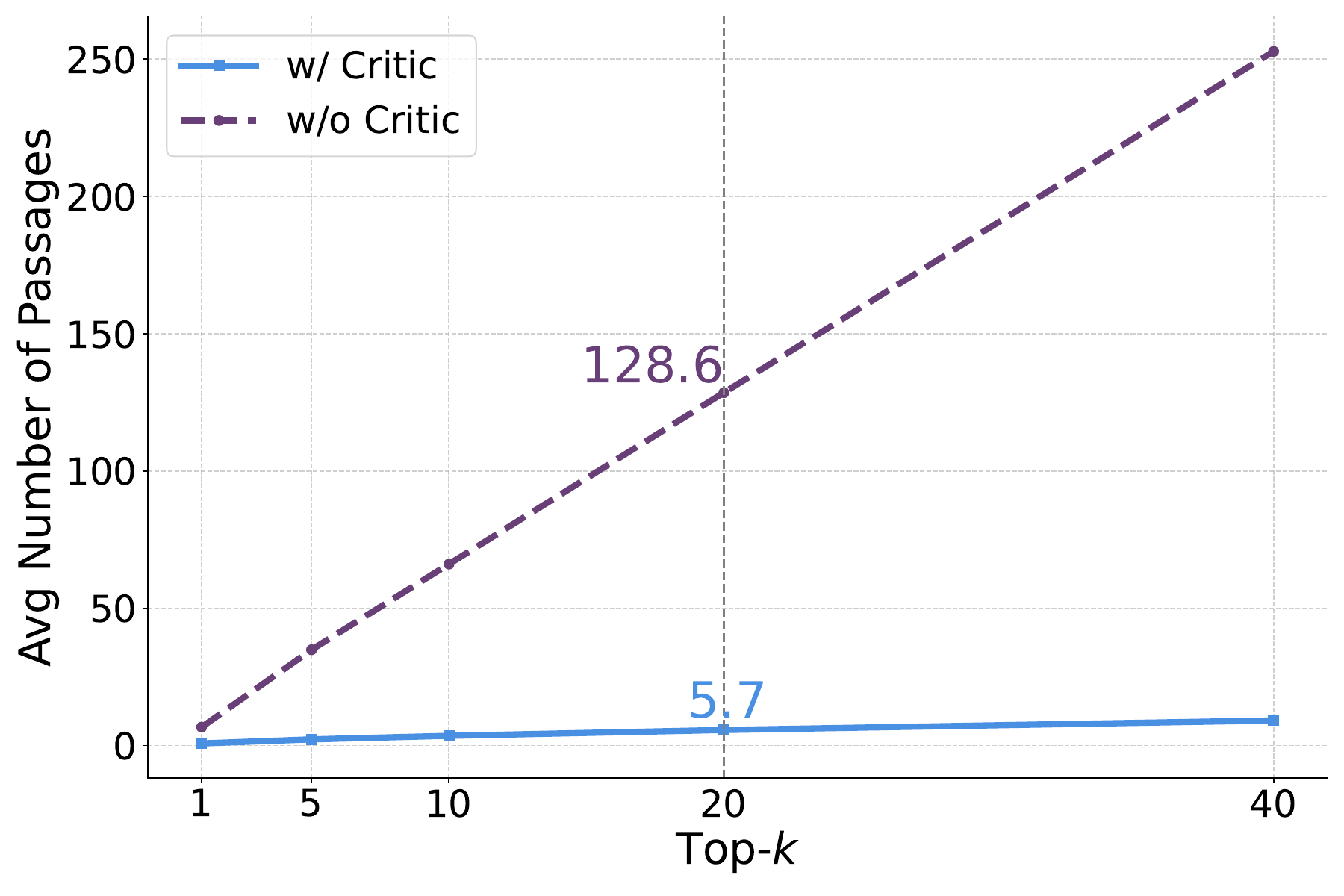}
    \vspace{-0.35cm}
    \caption{Comparison of the average number of passages fed to the generator with and without the critic filtering.
    }
    \label{fig:filtered_vs_unfiltered}
    \vspace{-0.4cm}
\end{figure}

\subsection{Effectiveness of the Critic Model} 

\tinytit{Critic Analysis} In Fig.~\ref{fig:filtered_vs_unfiltered}, we provide a detailed analysis on the effectiveness of the proposed critic model, employed in \ours to filter relevant passages. Specifically, the plot reports the average number of passages retained after the filtering performed by the critic model when varying the number $k$ of retrieved documents. Across all retrieval sizes, the critic model substantially reduces the number of retained passages (\eg, from an average of $128.6$ to $5.7$ at $k=20$), while preserving answer-relevant information. This highlights the strong ability of the critic model to discard noisy or off-topic passages, leading to a more compact and semantically aligned evidence set for multimodal reasoning.

\begin{figure}
   \centering
   \includegraphics[width=0.99\linewidth]{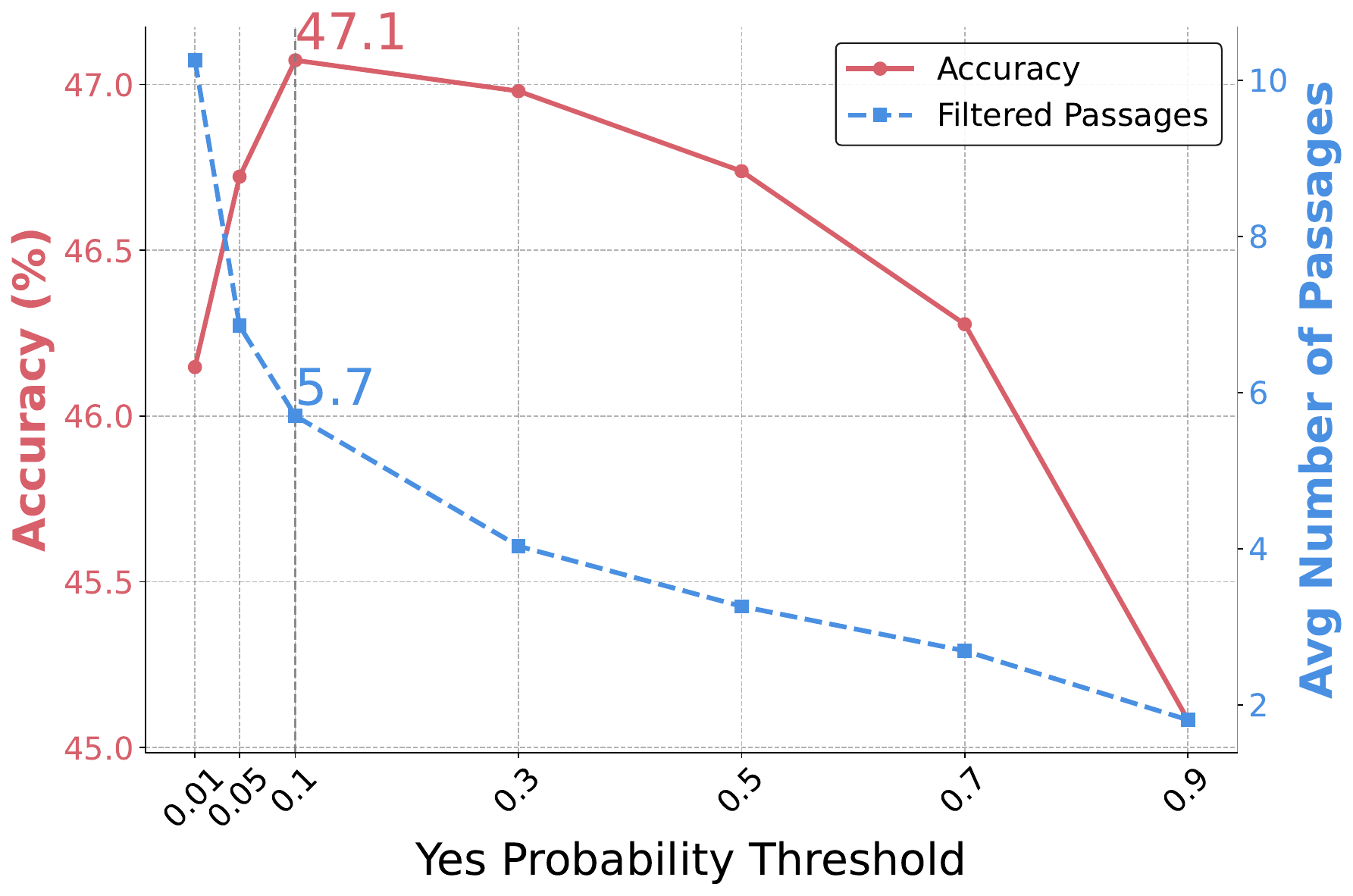}
   \vspace{-0.35cm}
    \caption{Performance of \ours 7B (red) and average number of filtered passages (blue) when varying the \emph{yes}-probability threshold in our critic model. Accuracy and number of passages are computed as the average of E-VQA and InfoSeek scores.}
    \label{fig:yes_prob_thr_analysis_avg}
    \vspace{-0.3cm}
\end{figure}

% \subsection{Critic Analysis}
\tit{Critic Threshold} In Fig.~\ref{fig:yes_prob_thr_analysis_avg}, we report how the performance varies as a function of the \emph{yes}-probability threshold used in our critic model (cf. Eq.~\ref{eq:critic2}). The results show that instead of simply letting the fine-tuned MLLM decide if the passage is relevant or not (\ie, $\text{thresh}=0.5$), leveraging the confidence of the model in predicting the ``Yes" token allows us to gain more control over the filtering phase. A threshold $\text{thresh}=0.1$ yields the best trade-off between precision and recall in retrieving relevant passages. This setting enables the critic model to reliably filter out passages for which it is most confident of their non-relevance to the query. 

% \section{Improvements Attribution}
% To better understand where the improvements of \ours originate, we analyze the contribution of the retrieval pipeline and of the generator training.

\tit{Filtering Analysis}
To evaluate the quality of our filtering stage, we report passage-level statistics measuring evidence retention (Recall) and irrelevant-passage recognition (Specificity) in Table~\ref{tab:critic_stats}. 
Our Critic retains relevant evidence with higher recall while filtering irrelevant passages with higher specificity compared to ReflectiVA. 
As a result, the generator receives a cleaner context with fewer passages on average, reducing noise while preserving useful evidence.

\subsection{Sources of Performance Gains}
% To better understand where the improvements of \ours originate, we analyze the contribution of the retrieval pipeline and of the generator training.

\tinytit{Impact of Evidence Quality}
As shown in the ablation study (Table~\ref{tab:ablation}), equipping the generator with our retrieval pipeline -- composed of the Critic and fine-grained retrieval -- already yields substantial gains in a Zero-Shot setting, reaching 40.2 on E-VQA and 27.1 on InfoSeek. 
This indicates that a significant portion of the improvements stems from higher-quality evidence selection rather than parametric memorization.

\tit{Training for Visual Robustness}
The SFT+RL procedure further improves performance by teaching the generator to leverage retrieved evidence when available and rely on visual reasoning otherwise. To encourage this behavior, the training data includes both relevant passages and distractors (30\% soft negatives and 70\% hard negatives), forcing the model to distinguish useful evidence from noise.

Finally, the relatively small differences in the \emph{w/o evidence} setting (Fig.~\ref{fig:graphs}, left) are partly due to a design choice in our retrieval pipeline. 
Unlike~\cite{cocchi2025augmenting}, which inserts a random passage when all evidence is filtered out, our method provides no fallback context, avoiding the risk of conditioning the generator on misleading information.

\begin{table}[t]
\caption{Recall and specificity for passage filtering on E-VQA.}
\vspace{-0.15cm}
\centering
\setlength{\tabcolsep}{.5em}
\resizebox{0.93\linewidth}{!}{
\begin{tabular}{l c c c c}
\toprule
% & & & \multicolumn{2}{c}{\textbf{Standard}} \\
% \cmidrule(lr){4-5}
\textbf{Model} &  & & Recall & Specificity \\
\midrule
ReflectiVA~\cite{cocchi2025augmenting} & LLaVA-MORE-8B & & 89.6 & 93.8 \\
ReflectiVA~\cite{cocchi2025augmenting} & Qwen2.5-VL-3B & & 91.3 & 93.8 \\
ReflectiVA~\cite{cocchi2025augmenting} & Qwen2.5-VL-7B & & 92.3 & 95.1 \\
\rowcolor{OurColor}
\textbf{\ours (Ours)} & Qwen2.5-VL-3B & & \textbf{94.6} & \textbf{95.7} \\
\bottomrule
\end{tabular}
}
\label{tab:critic_stats}
\vspace{-0.35cm}
\end{table}

\section{Critic and Generator Prompts}
\label{supp:prompts}
\tinytit{Critic Prompt Design}
The prompt illustrates the instruction used to query the critic model. For each image-question pair $(I_q, q)$ and a candidate passage $p$, the critic is explicitly asked to determine whether $p$ contains any information that could help answer the question. The formulation uses a minimal, binary response space (\ie, ``Yes''/``No''), which simplifies supervision and ensures consistent outputs across diverse samples. This concise design encourages the model to focus on relevance estimation rather than generative reasoning, enabling more stable fine-tuning and robust filtering of noisy passages during retrieval.

\begin{promptbox2}[title={Critic System Prompt}]
\footnotesize
You are a multimodal reasoning assistant specialized in Knowledge-Based Visual Question Answering (KB-VQA).

Your task is to evaluate whether a given text passage provides useful and relevant information for answering a question about an image.

You will be given:

- \textbf{Image}: a visual scene containing entities, actions, and context.

- \textbf{Question}: a natural-language question that refers to the image.

- \textbf{Text Passage}: an external knowledge snippet retrieved from a database.

You must analyze the semantic alignment between the text, the image, and the question.
Follow these steps carefully before giving your final decision:

1. Understand the visual scene: Identify the key objects, people, actions, and context visible in the image.

2. Interpret the question: Determine what information the question seeks (e.g., factual, reasoning, counting, attribute-based).

3. Analyze the text passage: Extract the main claims, facts, and entities mentioned in the text.

Compare for relevance: Assess whether the information in the text:

- Contains at least one sentence that supports answering the question about the image, OR

- Provides background knowledge needed to interpret or reason about the image-question pair.

Important:

- If even a single sentence in the passage is relevant or useful, consider the entire passage as relevant and answer "Yes”.

- If no part of the passage contributes meaningfully to answering the question, answer "No”.

Output only one word:

"\textbf{Yes}" -> if the text provides relevant or useful information for answering the question.

"\textbf{No}" -> if the text is irrelevant or unhelpful.
\end{promptbox2}

\begin{promptbox2}[title={Critic User Prompt}]
\footnotesize
Here is the question on the image above: 

\placeholder{Question}

Here is the text passage to analyze:
\placeholder{Passage}

Does the text passage contain at least one sentence that may have some information useful to answer the user question? 
"\textbf{Yes}"/"\textbf{No}" answer:
\end{promptbox2}

\tit{Generator Prompt Design}
This prompt defines the instruction for the generator model, which receives the image, question, and textual context. During training, the model is provided with only the single passage associated with the current example from the ReflectiVA dataset~\cite{cocchi2025augmenting}, whereas at inference it takes the subset of passages selected by the critic model. The generator is prompted to synthesize a final answer grounded in both visual and textual evidence. At inference, providing critic-filtered passages as input encourages concise, evidence-based reasoning, reduces the impact of irrelevant or noisy information, and improves factual grounding in multimodal responses. The generator system prompt is adapted from Dr. GRPO~\cite{liu2025understanding}.
Notably, when the critic model filters all retrieved passages (\ie, when $j=0$) the user prompt is changed and only the question with the image is fed to the generator.

\begin{promptbox1}[title={Generator System Prompt}]
\footnotesize
\texttt{A conversation between User and Assistant. The user asks a question, and the Assistant solves it. The assistant first thinks about the reasoning process and then provides the user with the answer. The reasoning process and answer are enclosed within <think> </think> and <answer> </answer> tags, respectively, i.e., <think>reasoning process here</think><answer>short answer here</answer>.}
\end{promptbox1}

\begin{promptbox1}[title={Generator User Prompt}]
\footnotesize
\placeholder{Question} 

The following paragraphs may contain useful information to help answer the question correctly:

<paragraph>\placeholder{$\text{Passage}_1$}</paragraph>

...

<paragraph>\placeholder{$\text{Passage}_j$}</paragraph>.
\end{promptbox1}

\tit{Reasoning-Trace Prompt Design}
To extract reasoning traces used during the SFT training stage (cf. Sec.~\ref{sec:cold_start} of the main paper), we employ a structured prompting strategy that elicits explicit, step-by-step inference from a teacher MLLM. The system prompt instructs the model to analyze the image, the question, and one retrieved passage, then produce a hidden reasoning trace ($\texttt{<think>} \dots \texttt{</think>}$) that (i) grounds its steps in visual evidence (\eg, objects, attributes, spatial relations), (ii) evaluates the content of the passage and explicitly states whether it is relevant or irrelevant, and (iii) connects visual and textual cues via a logical chain. The user prompt supplies the question, the retrieved passage with its relevance tag, and the correct answer; the model must output the trace plus the final answer in a strict schema.
Collected reasoning traces are used to initialize the generator with explicit reasoning trajectories that link the image, retrieved evidence, and the question, thereby strengthening its reasoning capabilities before the RL stage.

\begin{promptbox3}[title={Reasoning Traces Generation System Prompt}]
\footnotesize
You are a multimodal reasoning assistant.

Your goal is to analyze the image, the question, and the retrieved passage, and then produce a hidden reasoning trace that logically leads to the given answer.

The reasoning must be step-by-step, plausible, and based on both the visual evidence and the retrieved text passage.

You MUST explicitly state, within your reasoning trace, whether the passage is relevant or not according to the information provided (i.e., if it is labeled as "irrelevant", your reasoning must clearly and logically explain why it is not relevant, and if it is labeled as "relevant", your reasoning must logically support its relevance).

Do not mention, restate, or hint at the correct answer in the reasoning trace.

Your reasoning trace should include:

- Description of relevant visual evidence (objects, spatial relations, attributes).

- Analysis of the retrieved passage (what it states, whether it supports or contradicts the image/question, and its relevance).

- Logical deduction that connects the visual and textual evidence to reach a conclusion.

Here you have two good examples of reasoning traces:
\newline

EXAMPLE 1 with Relevant passage:

Question: \placeholder{Example 1 Question}

Retrieved \textbf{Relevant} Passage: 

\placeholder{Example 1 Relevant Passage}

Correct answer: \placeholder{Example 1 Answer}

Output:

<think> \placeholder{Example 1 Reasoning Trace} </think>

<answer> \placeholder{Example 1 Answer} </answer>
\newline

EXAMPLE 2 with Irrelevant passage:

Question: \placeholder{Example 2 Question}

Retrieved \textbf{Irrelevant} Passage: 

\placeholder{Example 2 Irrelevant Passage}

Correct answer: \placeholder{Example 2 Answer}

Output:

<think> \placeholder{Example 2 Reasoning Trace} </think>

<answer> \placeholder{Example 2 Answer} </answer>
\newline

Output your reasoning and the correct answer using the exact format below:

<think> [your reasoning trace here] </think>

<answer> [the provided answer] </answer>
\end{promptbox3}

\begin{promptbox3}[title={Reasoning Traces Generation User Prompt}]
\footnotesize
Question: \placeholder{Question}

Retrieved \textbf{\placeholder{Relevant}} Passage: \placeholder{Passage}

Correct Answer: \placeholder{Answer}

Please produce a reasoning trace that could logically lead to this answer, based on both the image and the retrieved passage if relevant. 

Do not mention or hint at the answer explicitly in your reasoning.

Concentrate on providing a coherent explanation that supports the indicated relevance or irrelevance of the passage in the reasoning trace, integrating both textual and visual evidence.

Make sure to insert the correct answer between the answer tags.
\end{promptbox3}

\section{Additional Qualitative Results}
\label{supp:qualitatives}

\begin{figure*}[t]
\centering
\small
\begin{minipage}{\linewidth}
    \begin{minipage}{0.3\linewidth}
        \includegraphics[width=1.\linewidth,height=0.88\linewidth]{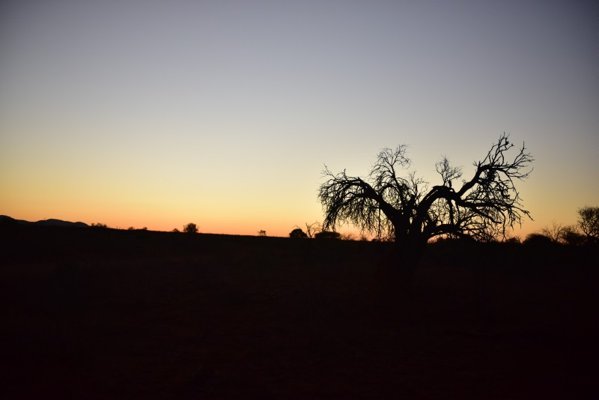}
    \end{minipage}
    \hspace{0.02\linewidth}
    \begin{minipage}{0.7\linewidth}
        {
        \textbf{Q}: Controversy surrounds the origin of what in the city from which one would drive to reach this nature reserve? 
        \vspace{0.15cm}
        
        \textbf{Qwen2.5-VL-7B (ZS)~\cite{bai2025qwen2}}:\\
        Based on the information provided, the city from which one would drive to reach the nature reserve is Port Augusta West. The text mentions that Port Augusta West is a suburb of Port Augusta, South Australia, and lists it as the starting point for driving to the nature reserve \textcolor{red}{\xmark} \\
        \textbf{ReflectiVA~\cite{cocchi2025augmenting}}:\\
        The Town \textcolor{red}{\xmark} \\
        \textbf{\ours (Ours):}\\
       \texttt{<think>} The image shows a nature reserve with a tree silhouette against a sunset sky, which matches the setting described in the retrieved passages. The question specifically asks for the origin of something in a city from which one would drive to reach this particular nature reserve. None of the retrieved passages provide any relevant information whatsoever regarding the question asked. Therefore, the final logical is that these passages must be disregarded insofar as the question is concerned. \texttt{</think>} \texttt{<answer>} Willi Stoph \textcolor[HTML]{00b050}{\cmark} \texttt{</answer>} 
        }
    \end{minipage}
    \vspace{0.3cm}
\end{minipage}

\begin{minipage}{\linewidth}
    \begin{minipage}{0.3\linewidth} 
        \includegraphics[width=1.\linewidth,height=0.88\linewidth]{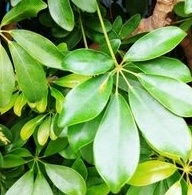}
    \end{minipage}
    \hspace{0.02\linewidth}
    \begin{minipage}{0.7\linewidth}
        {
        \textbf{Q}: What genus are this plant and schefflera actinophylla in? \vspace{0.15cm}
        
        \textbf{Qwen2.5-VL-7B (ZS)~\cite{bai2025qwen2}}:\\
        Not enough information is provided to determine the genus in question \textcolor{red}{\xmark} \\
        \textbf{ReflectiVA~\cite{cocchi2025augmenting}}:\\
        Schefflera \textcolor{red}{\xmark} \\
        \textbf{\ours (Ours):}\\
       \texttt{<think>} The image shows a plant that matches the description of Schefflera actinophylla in the retrieved passage. The question specifically asks for the genus of this particular plant in the image and Schefflera actinophylla. Both plants are identified as belonging to the same genus in the retrieved passage. The question directly corresponds to the information presented in the retrieved passage without any need to refer to the other retrieved passages. Therefore, the final verdict is that the final question must ignore these retrieved passages entirely. \texttt{</think>} \texttt{<answer>} Heptapleurum \textcolor[HTML]{00b050}{\cmark} \texttt{</answer>} 
        }
    \end{minipage}
    \vspace{0.9cm}
\end{minipage}

\begin{minipage}{\linewidth}
    \begin{minipage}{0.3\linewidth}
        \includegraphics[width=1.\linewidth,height=0.88\linewidth]{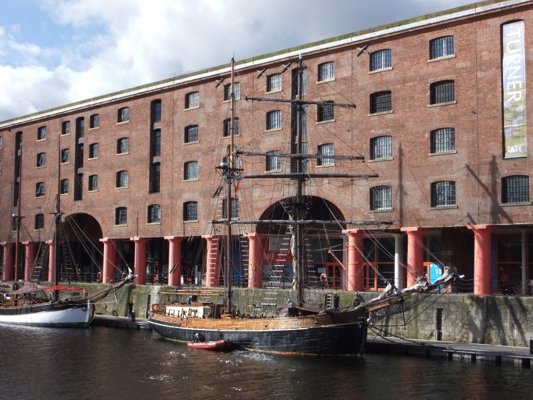}
    \end{minipage}
    \hspace{0.02\linewidth}
    \begin{minipage}{0.7\linewidth}
        {
        \textbf{Q}: Who designed this dock? \vspace{0.15cm}
        
        \textbf{Qwen2.5-VL-7B (ZS)~\cite{bai2025qwen2}}:\\
        Not enough information is provided to determine who designed the dock \textcolor{red}{\xmark} \\
        \textbf{ReflectiVA~\cite{cocchi2025augmenting}}:\\
        Jesse Hartley \textcolor{red}{\xmark} \\
        \textbf{\ours (Ours):}\\
       \texttt{<think>} The image shows the Albert Dock in Liverpool, which matches the question asked. The question specifically asks for the designer of this particular dock. The retrieved passage directly states that the dock in question, the Albert Dock, was designed by Jesse Hartley. Since the question specifically matches the dock in the image with the one mentioned in the passage, the retrieved passage directly answers the question without any need to refer to the question itself or the retrieved passage. \texttt{</think>} \texttt{<answer>} Royal Albert Dock \textcolor[HTML]{00b050}{\cmark} \texttt{</answer>} 
        }
    \end{minipage}
\end{minipage}
\vspace{-0.1cm}
\caption{Qualitative examples from Encyclopedic-VQA where we show the ability of \ours to reason about the question-image pair and the retrieved passages. For reference, we also report the generation from ReflectiVA~\cite{cocchi2025augmenting} and the corresponding zero-shot model.}
\label{fig:qualitatives_suppl_reasoning}
\end{figure*}

\tinytit{Reasoning Traces}
To further interpret the behavior of our model, we visualize qualitative examples of the reasoning traces generated by \ours in Fig.~\ref{fig:qualitatives_suppl_reasoning}. The zero-shot baseline produces partial reasoning but lacks a consistent structure and does not adhere to the output format defined by the evaluation datasets. In contrast, ReflectiVA follows the correct answer format but fails to generate explicit reasoning traces, limiting interpretability. In contrast, \ours generates coherent, well-structured traces that reveal the step-by-step logic behind its predictions. These examples highlight the ability of the proposed solution to integrate visual and textual cues, assess the relevance of retrieved passages, and maintain consistent reasoning even under noisy or irrelevant evidence, where baselines often over-rely on passages or hallucinate unsupported details.

\begin{figure*}[t]
\begin{minipage}{0.325\linewidth}
\scriptsize{\textbf{Q}: What did the zuni people use the bark of this \\ plant to dye black?\vspace{0.05cm}}\\
\begin{minipage}{0.443\linewidth}
\includegraphics[width=1.\linewidth,height=0.88\linewidth]{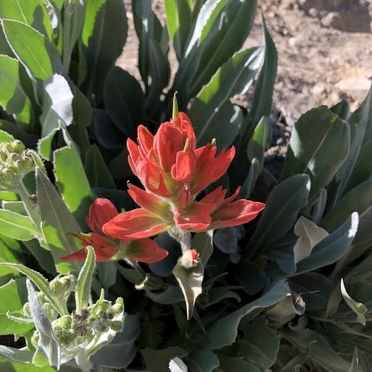}
\end{minipage}
\hfill
\begin{minipage}{0.53\linewidth}
\scriptsize{
\textbf{Qwen2.5-VL-7B (ZS)~\cite{bai2025qwen2}}:\\
The plant you are referring to is the Desert Paintbrush [..] \textcolor{red}{\xmark} \\
\textbf{ReflectiVA~\cite{cocchi2025augmenting}}:\\
Dyes \textcolor{red}{\xmark} \\
\textbf{\ours (Ours):}\\
Deerskin \textcolor[HTML]{00b050}{\cmark}
}
\end{minipage}
\end{minipage}
\hspace{0.02cm}
\begin{minipage}{0.325\linewidth}
\scriptsize{\textbf{Q}: What type of hawksbeard is this plant \\ commonly known as? \vspace{0.05cm}}\\
\begin{minipage}{0.443\linewidth}
\includegraphics[width=1.\linewidth,height=0.88\linewidth]{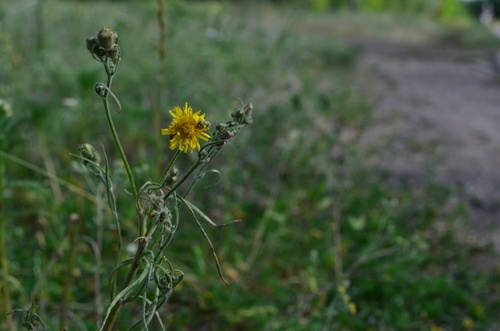}
\end{minipage}
\hfill
\begin{minipage}{0.53\linewidth}
\scriptsize{
\textbf{Qwen2.5-VL-7B (ZS)~\cite{bai2025qwen2}}:\\
The plant in the image is likely rough hawksbeard \textcolor{red}{\xmark} \\
\textbf{ReflectiVA~\cite{cocchi2025augmenting}}:\\
Mouse-ear \textcolor{red}{\xmark} \\
\textbf{\ours (Ours):}\\
Narrowleaf \textcolor[HTML]{00b050}{\cmark}
}
\end{minipage}
\end{minipage}
\hspace{0.02cm}
\vspace{0.1cm}
\begin{minipage}{0.325\linewidth}
\scriptsize{\textbf{Q}: What did Alexander Wilson do to this bird in 1811?\vspace{0.05cm}}\\
\begin{minipage}{0.443\linewidth}
\includegraphics[width=1.\linewidth,height=0.88\linewidth]{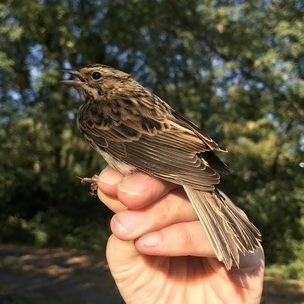}
\end{minipage}
\hfill
\begin{minipage}{0.53\linewidth}
\scriptsize{
\textbf{Qwen2.5-VL-7B (ZS)~\cite{bai2025qwen2}}:\\
Alexander Wilson did not harm the bird in any way [..] \textcolor{red}{\xmark} \\
\textbf{ReflectiVA~\cite{cocchi2025augmenting}}:\\
I don't see any information about a bird species \textcolor{red}{\xmark} \\
\textbf{\ours (Ours):}\\
Discovered \textcolor[HTML]{00b050}{\cmark}
}
\end{minipage}
\end{minipage}

\begin{minipage}{0.325\linewidth}
\scriptsize{\textbf{Q}: What part of a tree is this plant often found under? \vspace{0.05cm}}\\
\begin{minipage}{0.443\linewidth}
\includegraphics[width=1.\linewidth,height=0.88\linewidth]{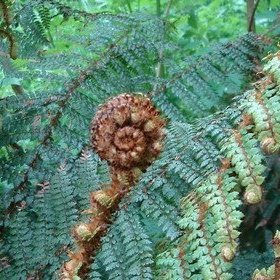}
\end{minipage}
\hfill
\begin{minipage}{0.53\linewidth}
\scriptsize{
\textbf{Qwen2.5-VL-7B (ZS)~\cite{bai2025qwen2}}:\\
Polystichum is often found under trees in disturbed habitats such as road cuts, talus slopes, and stream banks  \textcolor{red}{\xmark} \\
\textbf{ReflectiVA~\cite{cocchi2025augmenting}}:\\
Trunk \textcolor{red}{\xmark} \\
\textbf{\ours (Ours):}\\
Canopy \textcolor[HTML]{00b050}{\cmark}
}
\end{minipage}
\end{minipage}
\hspace{0.02cm}
\begin{minipage}{0.325\linewidth}
\scriptsize{\textbf{Q}: What monument was placed under state control in 1882? \vspace{0.05cm}}\\
\begin{minipage}{0.443\linewidth}
\includegraphics[width=1.\linewidth,height=0.88\linewidth]{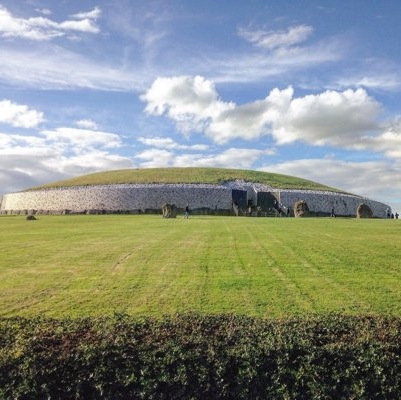}
\end{minipage}
\hfill
\begin{minipage}{0.53\linewidth}
\scriptsize{
\textbf{Qwen2.5-VL-7B (ZS)~\cite{bai2025qwen2}}:\\
Newgrange was placed under state control in 1882 \textcolor{red}{\xmark} \\
\textbf{ReflectiVA~\cite{cocchi2025augmenting}}:\\
Newgrange \textcolor{red}{\xmark} \\
\textbf{\ours (Ours):}\\
Newgrange and Knowth and Dowth \textcolor[HTML]{00b050}{\cmark}
}
\end{minipage}
\end{minipage}
\hspace{0.02cm}
\vspace{0.1cm}
\begin{minipage}{0.325\linewidth}
\scriptsize{\textbf{Q}: What type of habitat does this plant prefer?\vspace{0.05cm}}\\
\begin{minipage}{0.443\linewidth}
\includegraphics[width=1.\linewidth,height=0.88\linewidth]{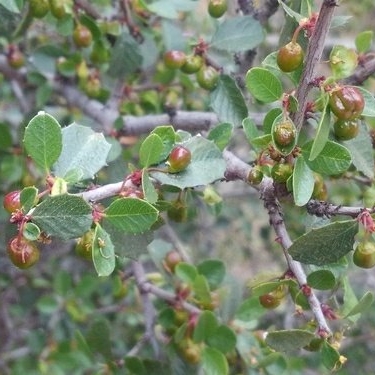}
\end{minipage}
\hfill
\begin{minipage}{0.53\linewidth}
\scriptsize{
\textbf{Qwen2.5-VL-7B (ZS)~\cite{bai2025qwen2}}:\\
It cannot be determined from the given information alone, but it is likely to [..] \textcolor{red}{\xmark} \\
\textbf{ReflectiVA~\cite{cocchi2025augmenting}}:\\
Dry \textcolor{red}{\xmark} \\
\textbf{\ours (Ours):}\\
Chaparral \textcolor[HTML]{00b050}{\cmark}
}
\end{minipage}
\end{minipage}

\begin{minipage}{0.325\linewidth}
\scriptsize{\textbf{Q}: In what german city is this landmark the most famous?\vspace{0.05cm}}\\
\begin{minipage}{0.443\linewidth}
\includegraphics[width=1.\linewidth,height=0.88\linewidth]{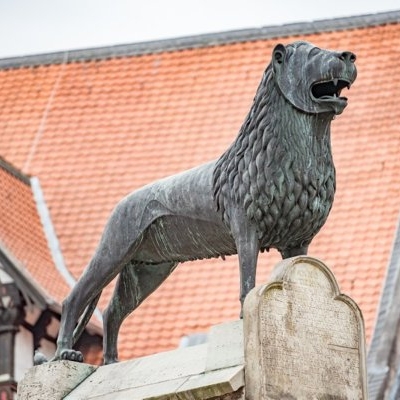}
\end{minipage}
\hfill
\begin{minipage}{0.53\linewidth}
\scriptsize{
\textbf{Qwen2.5-VL-7B (ZS)~\cite{bai2025qwen2}}:\\
The landmark in the picture is the Brunswick Lion \textcolor{red}{\xmark} \\
\textbf{ReflectiVA~\cite{cocchi2025augmenting}}:\\
Braunschweig \textcolor{red}{\xmark} \\
\textbf{\ours (Ours):}\\
Brunswick \textcolor[HTML]{00b050}{\cmark}
}
\end{minipage}
\end{minipage}
\hspace{0.02cm}
\begin{minipage}{0.325\linewidth}
\scriptsize{\textbf{Q}: What is the habitat of this animal? \vspace{0.05cm}}\\
\begin{minipage}{0.443\linewidth}
\includegraphics[width=1.\linewidth,height=0.88\linewidth]{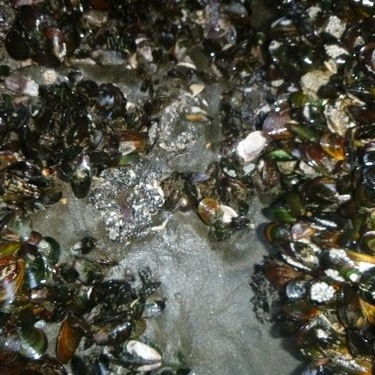}
\end{minipage}
\hfill
\begin{minipage}{0.53\linewidth}
\scriptsize{
\textbf{Qwen2.5-VL-7B (ZS)~\cite{bai2025qwen2}}:\\
In the upper intertidal zone on open coast [..] \textcolor{red}{\xmark} \\
\textbf{ReflectiVA~\cite{cocchi2025augmenting}}:\\
Salt marshes and quiet bays \textcolor{red}{\xmark} \\
\textbf{\ours (Ours):}\\
Intertidal zone on the open coast \textcolor[HTML]{00b050}{\cmark}
}
\end{minipage}
\end{minipage}
\hspace{0.02cm}
\vspace{0.1cm}
\begin{minipage}{0.325\linewidth}
\scriptsize{\textbf{Q}: Which teams played against each other in the first match ever held in this stadium? \vspace{0.05cm}}\\
\begin{minipage}{0.443\linewidth}
\includegraphics[width=1.\linewidth,height=0.88\linewidth]{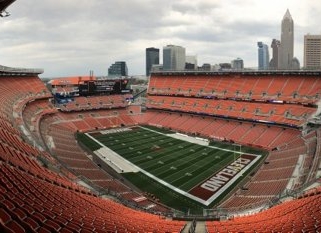}
\end{minipage}
\hfill
\begin{minipage}{0.53\linewidth}
\scriptsize{
\textbf{Qwen2.5-VL-7B (ZS)~\cite{bai2025qwen2}}:\\
The first match ever held in FirstEnergy Stadium, [..] \textcolor{red}{\xmark} \\
\textbf{ReflectiVA~\cite{cocchi2025augmenting}}:\\
Minnesota vikings \textcolor{red}{\xmark} \\
\textbf{\ours (Ours):}\\
Browns and Minnesota Vikings \textcolor[HTML]{00b050}{\cmark}
}
\end{minipage}
\end{minipage}

\begin{minipage}{0.325\linewidth}
\scriptsize{\textbf{Q}: What type of bird is the Contopus cooperi? \vspace{0.05cm}}\\
\begin{minipage}{0.443\linewidth}
\includegraphics[width=1.\linewidth,height=0.88\linewidth]{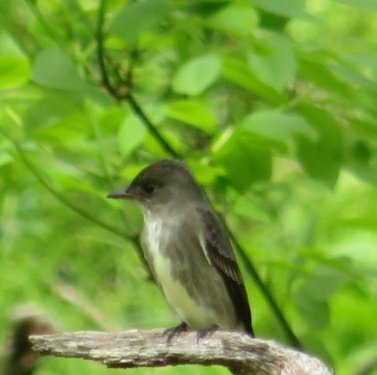}
\end{minipage}
\hfill
\begin{minipage}{0.53\linewidth}
\scriptsize{
\textbf{Qwen2.5-VL-7B (ZS)~\cite{bai2025qwen2}}:\\
Contopus cooperi (Olive-sided Flycatcher) \textcolor{red}{\xmark} \\
\textbf{ReflectiVA~\cite{cocchi2025augmenting}}:\\
Passerine \textcolor{red}{\xmark} \\
\textbf{\ours (Ours):}\\
Flycatcher \textcolor[HTML]{00b050}{\cmark}
}
\end{minipage}
\end{minipage}
\hspace{0.02cm}
\begin{minipage}{0.325\linewidth}
\scriptsize{\textbf{Q}: What month of meteorological winter in the northern hemisphere did hungary purchase this building? \vspace{0.05cm}}\\
\begin{minipage}{0.443\linewidth}
\includegraphics[width=1.\linewidth,height=0.88\linewidth]{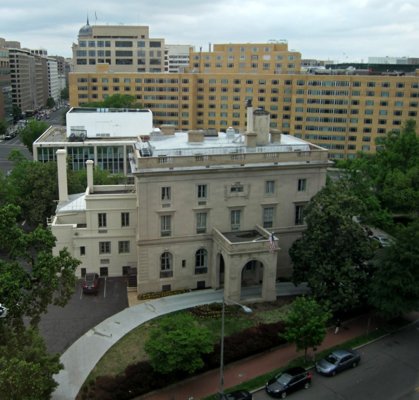}
\end{minipage}
\hfill
\begin{minipage}{0.53\linewidth}
\scriptsize{
\textbf{Qwen2.5-VL-7B (ZS)~\cite{bai2025qwen2}}:\\
According to the information provided, the U.S. Civil Service Building was built [..] in 1911 \textcolor{red}{\xmark} \\
\textbf{ReflectiVA~\cite{cocchi2025augmenting}}:\\
December \textcolor{red}{\xmark} \\
\textbf{\ours (Ours):}\\
Third \textcolor[HTML]{00b050}{\cmark}
}
\end{minipage}
\end{minipage}
\hspace{0.02cm}
\vspace{0.1cm}
\begin{minipage}{0.325\linewidth}
\scriptsize{\textbf{Q}: What wars ended the year the fichtelberg cable car was built on this mountain? \vspace{0.05cm}}\\
\begin{minipage}{0.443\linewidth}
\includegraphics[width=1.\linewidth,height=0.88\linewidth]{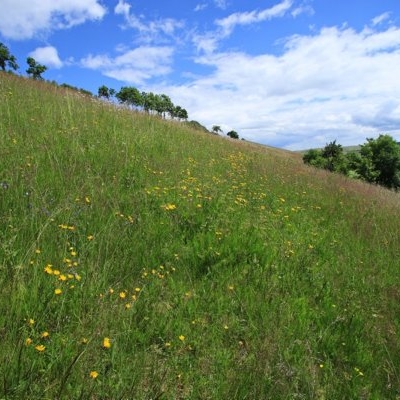}
\end{minipage}
\hfill
\begin{minipage}{0.53\linewidth}
\scriptsize{
\textbf{Qwen2.5-VL-7B (ZS)~\cite{bai2025qwen2}}:\\
Not enough information is provided to determine [..] \textcolor{red}{\xmark} \\
\textbf{ReflectiVA~\cite{cocchi2025augmenting}}:\\
World war i \textcolor{red}{\xmark} \\
\textbf{\ours (Ours):}\\
The American Indian Wars \textcolor[HTML]{00b050}{\cmark}
}
\end{minipage}
\end{minipage}

\begin{minipage}{0.325\linewidth}
\scriptsize{\textbf{Q}: Who dies without a male heir in the netherlands the year this plant was introduced to canada? \vspace{0.05cm}}\\
\begin{minipage}{0.443\linewidth}
\includegraphics[width=1.\linewidth,height=0.88\linewidth]{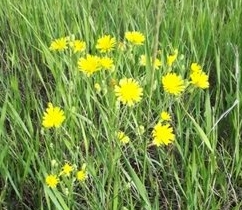}
\end{minipage}
\hfill
\begin{minipage}{0.53\linewidth}
\scriptsize{
\textbf{Qwen2.5-VL-7B (ZS)~\cite{bai2025qwen2}}:\\
Queen Beatrix \textcolor{red}{\xmark} \\
\textbf{ReflectiVA~\cite{cocchi2025augmenting}}:\\
The dutch royal family \textcolor{red}{\xmark} \\
\textbf{\ours (Ours):}\\
King William III \textcolor[HTML]{00b050}{\cmark}
}
\end{minipage}
\end{minipage}
\hspace{0.02cm}
\begin{minipage}{0.325\linewidth}
\scriptsize{\textbf{Q}: Who was the duke of milan when Edward Cowper began leading the monks at this abbey? \vspace{0.05cm}}\\
\begin{minipage}{0.443\linewidth}
\includegraphics[width=1.\linewidth,height=0.88\linewidth]{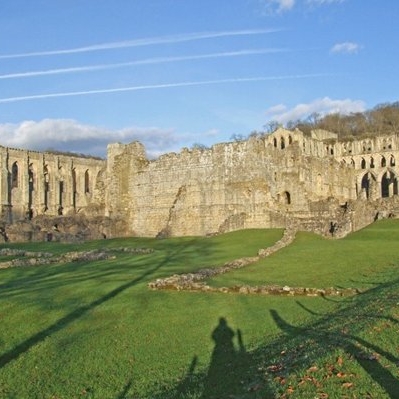}
\end{minipage}
\hfill
\begin{minipage}{0.53\linewidth}
\scriptsize{
\textbf{Qwen2.5-VL-7B (ZS)~\cite{bai2025qwen2}}:\\
The Duke of Brittany, William de Lis, was the owner of the patronage of Roche Abbey \textcolor{red}{\xmark} \\
\textbf{ReflectiVA~\cite{cocchi2025augmenting}}:\\
Charles i \textcolor{red}{\xmark} \\
\textbf{\ours (Ours):}\\
Maximilian Sforza \textcolor[HTML]{00b050}{\cmark}
}
\end{minipage}
\end{minipage}
\hspace{0.02cm}
\vspace{0.1cm}
\begin{minipage}{0.325\linewidth}
\scriptsize{\textbf{Q}: For what football cup final is this stadium the venue? \vspace{0.05cm}}\\
\begin{minipage}{0.443\linewidth}
\includegraphics[width=1.\linewidth,height=0.88\linewidth]{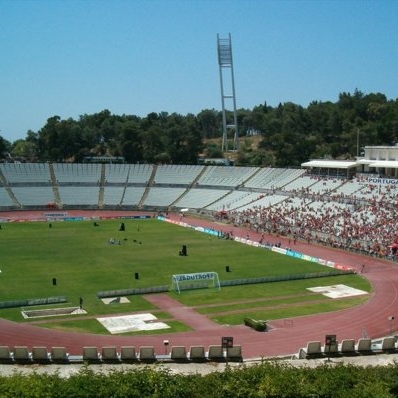}
\end{minipage}
\hfill
\begin{minipage}{0.53\linewidth}
\scriptsize{\textbf{Qwen2.5-VL-7B (ZS)~\cite{bai2025qwen2}}:\\
The football cup final for which this stadium is the venue is the Taça de Portugal \textcolor{red}{\xmark} \\
\textbf{ReflectiVA~\cite{cocchi2025augmenting}}:\\
Uefa women's championship \textcolor{red}{\xmark} \\
\textbf{\ours (Ours):}\\
Portuguese \textcolor[HTML]{00b050}{\cmark}
}
\end{minipage}
\end{minipage}

\begin{minipage}{0.325\linewidth}
\scriptsize{\textbf{Q}: In what month does it not snow in this lake? \vspace{0.05cm}}\\
\begin{minipage}{0.443\linewidth}
\includegraphics[width=1.\linewidth,height=0.88\linewidth]{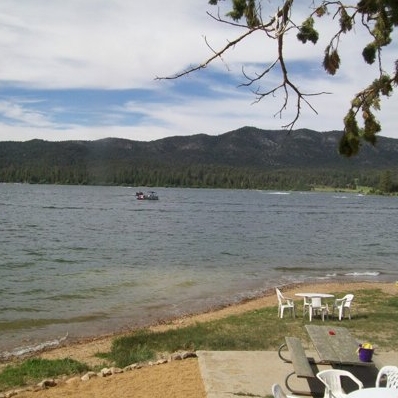}
\end{minipage}
\hfill
\begin{minipage}{0.53\linewidth}
\scriptsize{
\textbf{Qwen2.5-VL-7B (ZS)~\cite{bai2025qwen2}}:\\
The lake is not affected by snowfall, [..] \textcolor{red}{\xmark} \\
\textbf{ReflectiVA~\cite{cocchi2025augmenting}}:\\
July \textcolor{red}{\xmark} \\
\textbf{\ours (Ours):}\\
July and August \textcolor[HTML]{00b050}{\cmark}
}
\end{minipage}
\end{minipage}
\hspace{0.02cm}
\begin{minipage}{0.325\linewidth}
\scriptsize{\textbf{Q}: What is the religious order of this monastery? \vspace{0.05cm}}\\
\begin{minipage}{0.443\linewidth}
\includegraphics[width=1.\linewidth,height=0.88\linewidth]{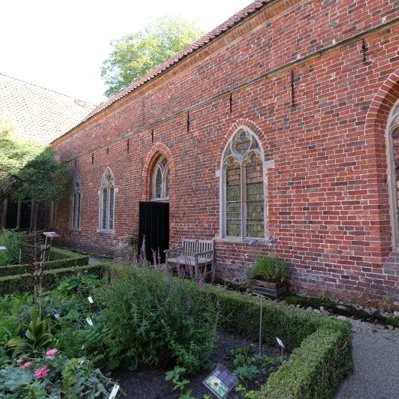}
\end{minipage}
\hfill
\begin{minipage}{0.53\linewidth}
\scriptsize{
\textbf{Qwen2.5-VL-7B (ZS)~\cite{bai2025qwen2}}:\\
Bremian Knighthood \textcolor{red}{\xmark} \\
\textbf{ReflectiVA~\cite{cocchi2025augmenting}}:\\
Premonstratensian \textcolor{red}{\xmark} \\
\textbf{\ours (Ours):}\\
 Order of the Holy Cross \textcolor[HTML]{00b050}{\cmark}
}
\end{minipage}
\end{minipage}
\hspace{0.02cm}
\vspace{0.1cm}
\begin{minipage}{0.325\linewidth}
\scriptsize{\textbf{Q}: What is the surface area of this lake? \vspace{0.05cm}}\\
\begin{minipage}{0.443\linewidth}
\includegraphics[width=1.\linewidth,height=0.88\linewidth]{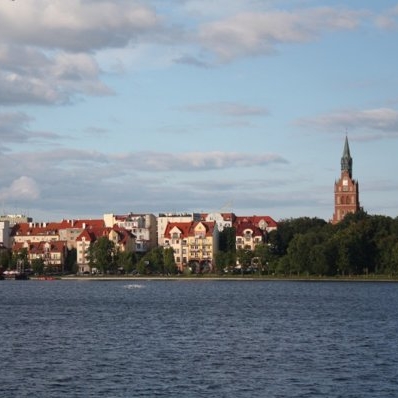}
\end{minipage}
\hfill
\begin{minipage}{0.53\linewidth}
\scriptsize{
\textbf{Qwen2.5-VL-7B (ZS)~\cite{bai2025qwen2}}:\\
Ełk Lake is a fresh water lake in the Masurian Lake District of Poland's Warmia-Mazury Province \textcolor{red}{\xmark} \\
\textbf{ReflectiVA~\cite{cocchi2025augmenting}}:\\
3.82 \textcolor{red}{\xmark} \\
\textbf{\ours (Ours):}\\
3.82 square kilometres \textcolor[HTML]{00b050}{\cmark}
}
\end{minipage}
\end{minipage}

\begin{minipage}{0.325\linewidth}
\scriptsize{\textbf{Q}: Who designed this palace?
 \vspace{0.05cm}}\\
\begin{minipage}{0.443\linewidth}
\includegraphics[width=1.\linewidth,height=0.88\linewidth]{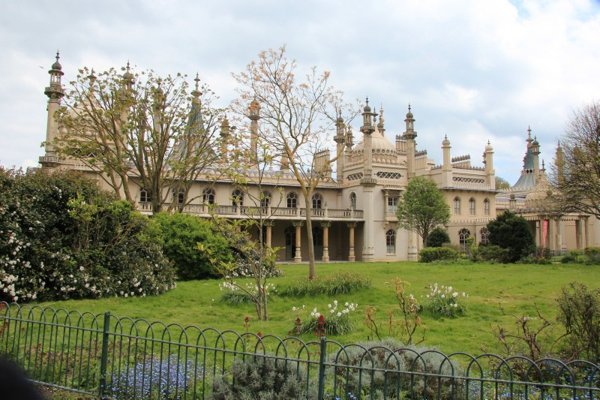}
\end{minipage}
\hfill
\begin{minipage}{0.53\linewidth}
\scriptsize{
\textbf{Qwen2.5-VL-7B (ZS)~\cite{bai2025qwen2}}:\\
Not enough information is provided to determine [..]  \textcolor{red}{\xmark} \\
\textbf{ReflectiVA~\cite{cocchi2025augmenting}}:\\
Peter Frederick Robinson \textcolor{red}{\xmark} \\
\textbf{\ours (Ours):}\\
John Nash \textcolor[HTML]{00b050}{\cmark}
}
\end{minipage}
\end{minipage}
\hspace{0.02cm}
\begin{minipage}{0.325\linewidth}
\scriptsize{\textbf{Q}: In which country or region does this animal live?
 \vspace{0.05cm}}\\
\begin{minipage}{0.443\linewidth}
\includegraphics[width=1.\linewidth,height=0.88\linewidth]{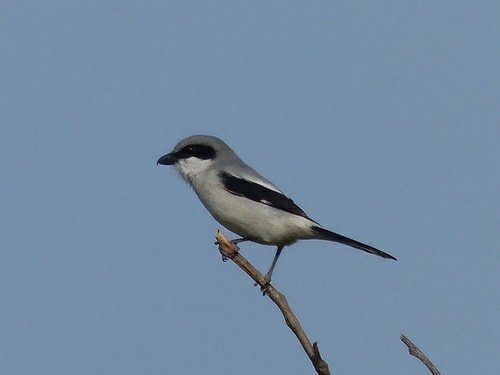}
\end{minipage}
\hfill
\begin{minipage}{0.53\linewidth}
\scriptsize{
\textbf{Qwen2.5-VL-7B (ZS)~\cite{bai2025qwen2}}:\\
It is native to southern Canada, the contiguous USA, and Mexico \textcolor{red}{\xmark} \\
\textbf{ReflectiVA~\cite{cocchi2025augmenting}}:\\
Eurasia \textcolor{red}{\xmark} \\
\textbf{\ours (Ours):}\\
North America \textcolor[HTML]{00b050}{\cmark}
}
\end{minipage}
\end{minipage}
\hspace{0.02cm}
\vspace{0.1cm}
\begin{minipage}{0.325\linewidth}
\scriptsize{\textbf{Q}: How many eggs does this bird typically lay? \vspace{0.05cm}}\\
\begin{minipage}{0.443\linewidth}
\includegraphics[width=1.\linewidth,height=0.88\linewidth]{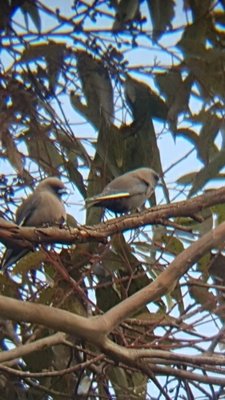}
\end{minipage}
\hfill
\begin{minipage}{0.53\linewidth}
\scriptsize{
\textbf{Qwen2.5-VL-7B (ZS)~\cite{bai2025qwen2}}:\\
Three eggs \textcolor{red}{\xmark} \\
\textbf{ReflectiVA~\cite{cocchi2025augmenting}}:\\
Three \textcolor{red}{\xmark} \\
\textbf{\ours (Ours):}\\
Three to four \textcolor[HTML]{00b050}{\cmark}
}
\end{minipage}
\end{minipage}

\vspace{-0.25cm}
\caption{Qualitative results on InfoSeek and Encyclopedic-VQA image-question pairs comparing \ours, ReflectiVA~\cite{cocchi2025augmenting}, and the corresponding zero-shot model.}
\label{fig:qualitatives_suppl}
\vspace{-0.35cm}
\end{figure*}

\tit{KB-VQA Qualitative Results}
Fig.~\ref{fig:qualitatives_suppl} presents additional qualitative examples from the InfoSeek and Encyclopedic-VQA benchmarks, comparing the responses of \ours, ReflectiVA~\cite{cocchi2025augmenting}. As shown, \ours produces answers that remain aligned with both the visual content and the retrieved evidence, benefiting from its critic-guided filtering and structured reasoning. These examples further highlight the robustness of \ours in handling complex, knowledge-driven VQA scenarios.

\section{Limitations and Impact} 
\label{supp:limitations}
While \ours demonstrates strong performance across standard benchmarks, it still faces some limitations. First, the generator produces a detailed reasoning trace, which improves the explainability of the final answer but may also increase latency, as more tokens must be generated before producing the answer. Second, the quality of \ours depends on the reliability of the retrieved evidence. Although the critic effectively filters irrelevant passages, retrieval failures or missing knowledge can still lead to incomplete or incorrect reasoning. Moreover, the model may occasionally over-structure its explanations, producing reasoning that is correct in format but not perfectly aligned with human logic. 

Despite these limitations, the explicit separation of evidence filtering from reasoning and answer generation enables \ours to achieve strong performance while promoting greater transparency and explainability, potentially inspiring future research on modular and trustworthy multimodal reasoning frameworks.

\end{document}